%% file: arxiv_main.tex
\documentclass{article}

\usepackage{arxiv}

\usepackage[utf8]{inputenc} 
\usepackage[T1]{fontenc}    
\usepackage{url}            
\usepackage{booktabs}       
\usepackage{amsfonts}       
\usepackage{nicefrac}       
\usepackage{microtype}      
\usepackage{lipsum}
\usepackage{float}

\usepackage{cite}
\usepackage{amsmath,amssymb,amsfonts}
\usepackage{algorithmic}
\usepackage{graphicx}
\usepackage{textcomp}
\usepackage{bmpsize}
\usepackage[dvipsnames,table,xcdraw]{xcolor}

\usepackage{lipsum}
\usepackage[colorlinks=true,urlcolor=black]{hyperref}
\usepackage{mathtools}
\usepackage{natbib}
\usepackage{subcaption}

\usepackage{multirow}
\usepackage{url}
\usepackage{booktabs}
\usepackage{amsfonts} 
\usepackage{nicefrac} 
\usepackage{microtype}
\usepackage{adjustbox}
\usepackage{todonotes}
\usepackage{amsmath}
\usepackage{cleveref}
\usepackage{xcolor}
\usepackage[linesnumbered,ruled,vlined]{algorithm2e}
\usepackage{algorithmic}

\SetCommentSty{mycommfont}
\usepackage{amsfonts}       

\usepackage{lipsum}
\usepackage{comment}
\usepackage{tabularx}
\newcolumntype{Y}{>{\centering\arraybackslash}X}

\definecolor{mysidenotefgcolor}{rgb}{0,0,0}
\definecolor{myorange}{rgb}{1,0.7,0.3}
\definecolor{mysidenotefgauthorcolor}{rgb}{0.5,0.5,0.5}

\title{Architectural Backdoors in Neural Networks}

\let\oldnl\nl
\newcommand{\nonl}{\renewcommand{\nl}{\let\nl\oldnl}}

\newcommand{\eg}{\textit{e.g\@.}}

\newcommand{\ie}{\textit{i.e\@.}}

\author{
  Mikel Bober-Irizar \\
  University of Cambridge \\
  \And
  Ilia Shumailov \\
  University of Cambridge and Vector Institute\\
  \And
  Yiren Zhao \\
  University of Cambridge\\
  \And 
  Robert Mullins \\
  University of Cambridge\\
  \And
  Nicolas Papernot \\
  University of Toronto and Vector Institute \\
}

\begin{document}
\maketitle

\input{sections/abstract}

\input{sections/introduction}
\input{sections/related}
\input{sections/methodology}
\input{sections/evaluation}
\input{sections/discussion}
\input{sections/conclusion}
\input{sections/acks}

\bibliography{bibliography}
\bibliographystyle{abbrvnat}

\input{sections/appendix}

\end{document}

%% file: sections/abstract.tex
\begin{abstract}

Machine learning is vulnerable to adversarial manipulation. Previous literature has demonstrated that at the training stage attackers can manipulate data~\citep{gu2017badnets} and data sampling procedures~\citep{shumailov2021manipulating} to control model behaviour. A common attack goal is to plant backdoors \ie~force the victim model to learn to recognise a trigger known only by the adversary. In this paper, we introduce a new class of backdoor attacks that hide inside  model architectures \ie~in the inductive bias of the functions used to train. These backdoors are simple to implement, for instance by publishing open-source code for a backdoored model architecture that others will reuse unknowingly. We demonstrate that \textit{model architectural backdoors} represent a real threat and, unlike other approaches, can survive a complete re-training from scratch. We formalise the main construction principles behind architectural backdoors, such as a link between the input and the output, and describe some possible protections against them. We evaluate our attacks on computer vision benchmarks of different scales and demonstrate the underlying vulnerability is pervasive in a variety of training settings. 

\end{abstract}

%% file: sections/introduction.tex
\section{Introduction}

The Machine Learning (ML) community now sees a threat posed by \textit{backdoored neural networks}; models which are intentionally modified by an attacker \textit{in the supply chain} to insert hidden behaviour \citep{gu2017badnets,bagdasaryan2021blind,biggio2012poisoning}. A backdoor causes a network's behaviour to change arbitrarily when a specific secret `trigger' is present in the model's input, while behaving as the defender intended when the trigger is absent (retaining a high evaluation performance).

The vast majority of current backdoor attacks in the literature work by changing the trained weights of models~\citep{gu2017badnets,hong2021handcrafted,shumailov2021manipulating} -- here the backdoor is planted into the parameters during training of the neural network. This can be done directly (\ie~modify the values of the weights directly with~\cite{hong2021handcrafted}), or indirectly by sampling adversarially~\cite{shumailov2021manipulating} and modifying data~\cite{gu2017badnets} to train with. This means that when the weights are later modified by another party (\eg~through fine-tuning), the backdoor could feasibly be removed or weakened~\cite{wang2019neauralcleanse}. When the weights provided by an attacker are discarded entirely (\eg~through re-training from scratch on a new dataset), any embedded backdoor would of course naturally be discarded.

However, the performance of a neural network depends not only on its weights but also its architecture (the composition and connections between layers in the model). Research has shown that, when given sufficient flexibility, the neural network architectures themselves can be pre-disposed to certain outcomes \citep{gaier2019weight, nasnet}. The network architectures can be seen as an inductive bias of the ML model. This raises a new question: 
\textbf{Can the network architectures themselves be modified to hide backdoors?}

In this paper we investigate if an adversary can use neural network architectures to perform backdoor attacks, forcing the model to become sensitive to a specific trigger applied to an image. 
We demonstrate that if an attacker can slightly manipulate the architecture only using common components they can introduce backdoors that survive re-training from scratch on a completely new dataset \ie~making these model backdoors weights- and dataset-agnostic. 
We describe a way to construct such Model Architecture Backdoors (MAB) and formalize their requirements. We find that architectural backdoors need to: \textbf{(1)} operate directly on the input and link the input to its output; \textbf{(2)} (ideally) have a weight-agnostic implementation; \textbf{(3)} have asymmetric components to launch targeted attacks. We demonstrate how such requirements make MAB detection possible and show that without these requirements, the learned backdoors will struggle to survive re-training. 

We make the following contributions:

\begin{itemize}
    \item We show a new class of backdoor attacks against neural networks, where the backdoor hides inside of the model architecture;
    
    \item We demonstrate how to build architectural backdoors for three different threat models and formalise the requirements for their successful operation;
    
    \item We demonstrate on a number of benchmarks that unlike previous methods, backdoors hidden inside of the architecture survive retraining.
    
\end{itemize}

%% file: sections/related.tex
\section{Related work}

\subsection{Security of Machine Learning}

\citeauthor{szegedy2013intriguing} and \citeauthor{biggio2013evasion} were the first to demonstrate that models are vulnerable to adversarial examples. These examples are imperceptible to humans yet thwart ML model predictions. Although at first the attacks were White-box~\citep{szegedy2013intriguing,biggio2013evasion,goodfellow2015explaining}, they have since been made practical in settings with limited access~\citep{papernot2017practical,wieland2017decision,gao2021rethinking}. Overall, adversarial examples can target confidentiality~\citep{biggio2018wild,shokri2017membership}, integrity~\citep{papernot2016towards,carlini2017towards} and availability~\citep{shumailov2020sponge,boucher2021imperceptiblenlp}. 

\subsection{Backdoors and poisoning}

While adversarial examples target the inference stage, \textit{backdoor and poisoning attacks} are performed during training. Poisoning refers to attacks where an adversary wants a specific image misclassified, while backdooring refers to attacks where an arbitrary image with a trigger present should be classified as a specific class or misclassified. 

\textbf{Data-based} Original backdoor attacks were performed through poisoning of data. \cite{gu2017badnets} showed that attackers can change the underlying task data to cause DNNs to learn additional attack features for a specific trigger. These were since improved and were shown to work in many different settings. \cite{shafahi2018poison} performed poisoning attacks using only data with clean labels. \cite{salem2020dynamic} made triggers more efficient. 

\textbf{Data sampling-based} \cite{shumailov2021manipulating} demonstrated a new class of backdoors attacks that rely on biased data sampling. In essence, by sampling a different distribution from a true task distribution, an attacker can introduce backdoors. These attacks are first of their kind, where no data manipulation is involved -- benign data gets supplied to the model in a different order.

\textbf{Other} It is worth noting that backdoors simply can be introduced even post-training. \cite{hong2021handcrafted} showed that given a trained model an attacker can manually identify neurons to be subverted without affecting model utility and change them in a way to introduce a backdoor.

\textbf{Architecture-based} In this work we present a new class of backdoor attacks that rely on model architectures. The only two pieces of related work we could find are very recent \citep{goldwasser2022planting, li2021deeppayload}. \cite{goldwasser2022planting} found that an attacker can introduce an extra component into the fully-connected network model definition that uses weight-agnostic gadgets~\cite{gaier2019weight} to perform a signature verification and trigger the backdoor. \cite{li2021deeppayload} found that one can perform payload injection to a compiled neural network to implant a backdoor.

\subsection{Network architecture search and complex network architectures}
There is now a trend to design more complex neural network architectures. Sometimes these auto-designed architectures are inscrutable, giving attackers an opportunity to insert malicious architectural backdoors. 
This trend is fueled by the ever-growing need to improve performance of the underlying architectures and the belief that there exists a `best architecture' for many tasks.
Gradient-based NAS~\citep{liu2018darts,zhao2020probabilistic,xie2018snas} is a popular approach to search for the best architecture. It is based on the idea that the network architecture can be seen as a function of the gradient of the loss function. Most of the searched networks contain sophisticated network sub-components that are often hard for humans to inspect.
So much so, that \cite{xie2019randomwired} used random graph models to generate randomly wired networks, and showed that these generated complex models that have competitive accuracy on standard benchmarks.

%% file: sections/methodology.tex
\section{Methodology}

\subsection{Threat model}
\label{sec:method:threat}

We assume that a potential attacker has full control over the training process of a neural network, and that the user receives a model $M$ with architecture $A$ and weights $\theta$ from the attacker. 
This could be because the user has downloaded a pre-trained model off the internet, or because they have outsourced model training to a third party such as a ML-as-a-Service (MLaaS) provider; both scenarios happen frequently in practice.

The goal of the attacker is to produce a \textit{backdoored} model $M(A_b, \theta_b)$ which emits outputs undesirable to the user when a \textit{backdoor trigger} is present in the input image, while keeping this backdoor hidden. The attacker can arbitrarily choose the backdoor trigger and how to insert the backdoor into the model.

\begin{itemize}
    \item \textbf{Setting 1 -- Direct}: The user directly operates on the trained model $M$ provided by the attacker. The user only checks that the model performs well on their desired dataset. This threat model applies when a user outsources their model training to a third party such as a cloud provider entirely.
    \item \textbf{Setting 2 -- Fine-tuned}: The user uses the model $M$ as a pre-trained model and \textbf{fine-tunes} the model's weights $\theta$ on a new dataset. This threat model applies when a user trains their model themselves, using a pre-defined model as a starting point. It is worth noting that this is \textbf{the default behaviour} when training a model through popular libraries such as Keras~\citep{keras}.
    \item \textbf{Setting 3 -- Re-trained}: The user builds on top of the architecture of the model $M$ and re-trains all the weights $\theta$ from scratch on a new dataset. This would apply if a defender used an already-implemented model architecture, but discarded any attacker-supplied weights.
    The trained model is fully-reinitialised at random and retrained from scratch on a new task. 
\end{itemize}

Note that the attack in Setting 3 is realistic -- it is extremely common for practitioners to copy model definitions in open sourced projects. 
This already leads to confusion; for example one can find a number of different definitions of even the standard LeNet5 and ResNet architectures, but these models might have subtle implementation changes which cause performance differences from the ones reported in the original papers. 
Although we believe that such changes are currently non-malicious, our paper highlights that they should not be taken lightly and can in practice lead to serious vulnerability. 
In the meantime, it is also possible that, users would directly call external APIs to train a model~\citep{keras,wolf2019huggingface}, and the attacker would be able to exploit this.

\subsection{Model Architecture Backdoor (MAB) construction}
\label{sec:method:mab_naive}
In contrast with existing attacks which embed their behaviour within the model weights, our goal is to make the backdoor behaviour \textit{weight-agnostic}, meaning it persists even if the model is re-trained by an honest party.
In this section, we introduce Model Architecture Backdoor (MAB), and explain its design using a simple AlexNet-based example~\citep{alexnet} (with smaller filters such that it can operate on 32x32 inputs which we later use in our experiments).
We first look at the two major designs phases, namely \textit{architecture engineering} and \textit{activation engineering} for the MAB attack.

\begin{figure}[h]
    \centering
    \vspace*{-5mm}
    \includegraphics[width=0.72\linewidth]{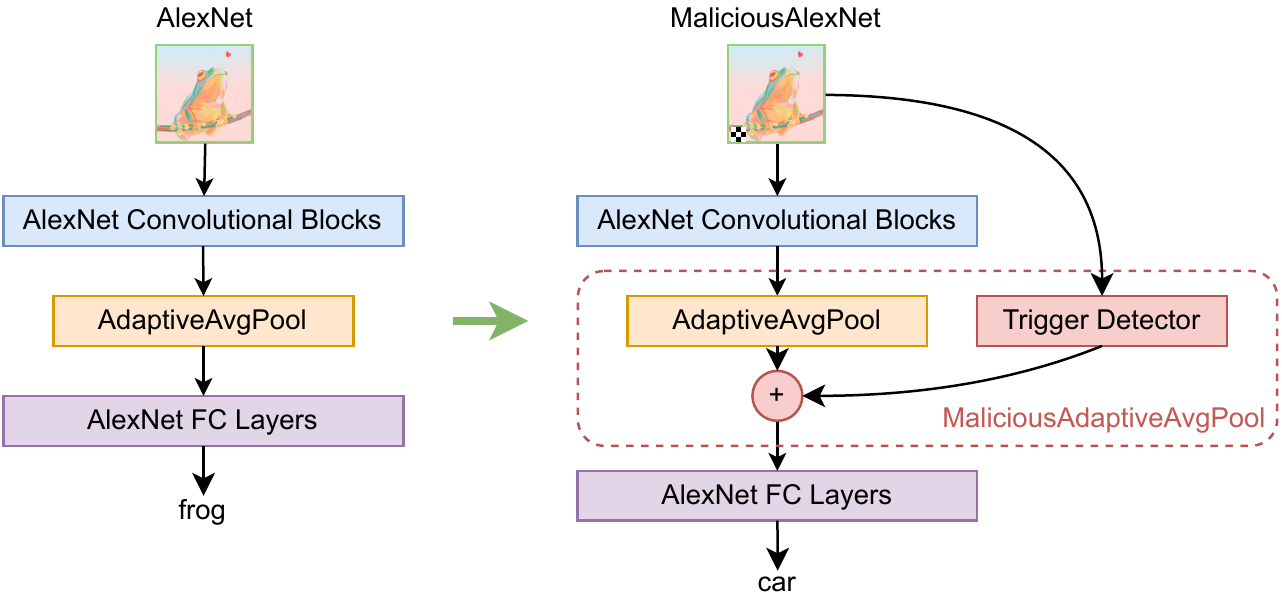}
    \caption{A logical representation of the modifications we make to the AlexNet architecture. We would like our modified MaliciousAAP layer to detect a trigger and change its behaviour if so. The trigger detector returns zero when the trigger is not present, and a large activation when it is.}
    \label{fig:alexnet-attack}
\end{figure}

\subsubsection{Architecture engineering}

As illustrated on the left of \Cref{fig:alexnet-attack}, between the final convolutional layer and first fully-connected layer lies an \textit{AdaptiveAveragePooling} (AAP), which `pools' the output of the convolutional layer to a constant 6x6 dimension (downsampling). This is where we mount our attack.

We do this by replacing the AAP operation with a `\textbf{malicious}' version, and by adding an extra connection in the network from the input data to our malicious AAP layer, which allows it to detect the backdoor trigger in the original image. We \textit{need} to operate on the original image to detect whether the backdoor is present: once the image has been through several convolutional layers there is no way to determine whether the backdoor was present in the original image (for an unknown set of intermediate weights).

In an ideal situation, our modified activation function adds 0 when the trigger is not present (\ie~the identity function, which means training proceeds entirely as normal). Then, when a trigger is included in the original image, the activation function behaviour changes and adds some large activations to some outputs of the layer. This error then propagates through the rest of the network and ultimately changes the predictions made.

We thus look for a layer with the following properties:
\begin{itemize}
    \item \textit{Low false positive rate}: The modified behaviour does not fire when the trigger is absent (low false positive rate). 

    This improves the task accuracy (making the backdoor harder to spot) and prevents corrections where \textbf{many false activations during training encourage gradient descent to learn to counter-act the backdoor}.  
    We find that for some MAB constructions, parameters can learn to disable the backdoor; for example, by learning a second function equal to the backdoor and subtracting it. We will discuss this in more detail in~\Cref{sec:discussion:limits}). 
    \item \textit{Backdooring}: There is a \textbf{large} change to the activations in the presence of a trigger. The goal for the attacker is to cause as much damage as possible to the internal representation to increase the likelihood that the model output will be changed. Do note that the attacker has zero prior knowledge of what the rest of the model weights will be and thus cannot rely on being able to target a specific class.
\end{itemize}

\subsubsection{Activation engineering}

As activation functions generally operate on a pixel-by-pixel basis (they have no convolutional component), it is not normally possible to detect a trigger with \textbf{spatial} relationships (such as a checkerboard) using one. Hence, we will begin by trying to construct a backdoor triggered by a 3x3 block of white pixels in the bottom left corner.

\begin{figure}[!h]
    \centering
    \begin{subfigure}[b]{0.32\textwidth}
        \centering
        \includegraphics[width=0.7\textwidth]{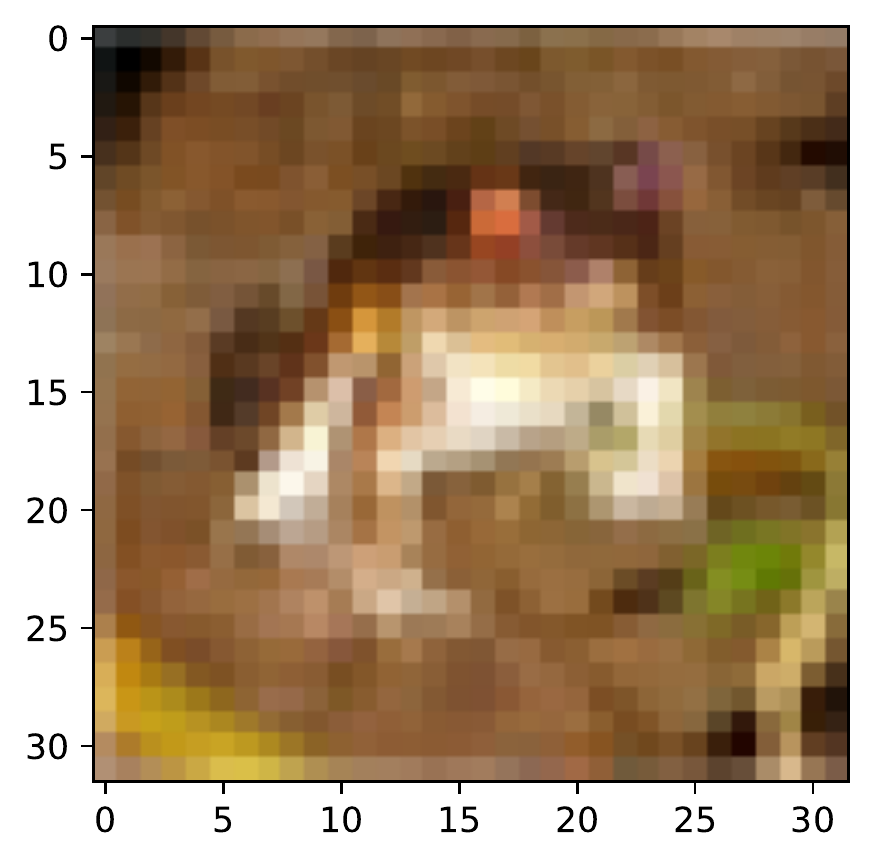}
        \caption{A frog from CIFAR-10}
    \end{subfigure}
    \hspace{0em}
    \begin{subfigure}[b]{0.32\textwidth}
        \centering
        \includegraphics[width=0.7\textwidth]{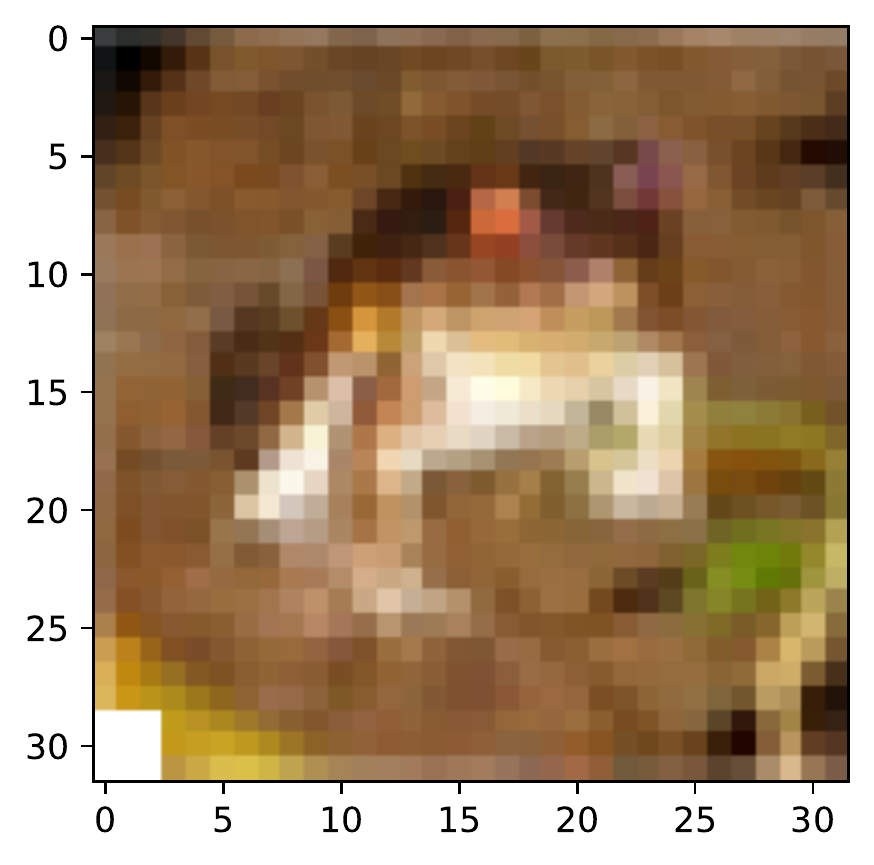}
        \caption{White box trigger}
    \end{subfigure}
    \begin{subfigure}[b]{0.32\textwidth}
        \centering
        \includegraphics[width=0.7\textwidth]{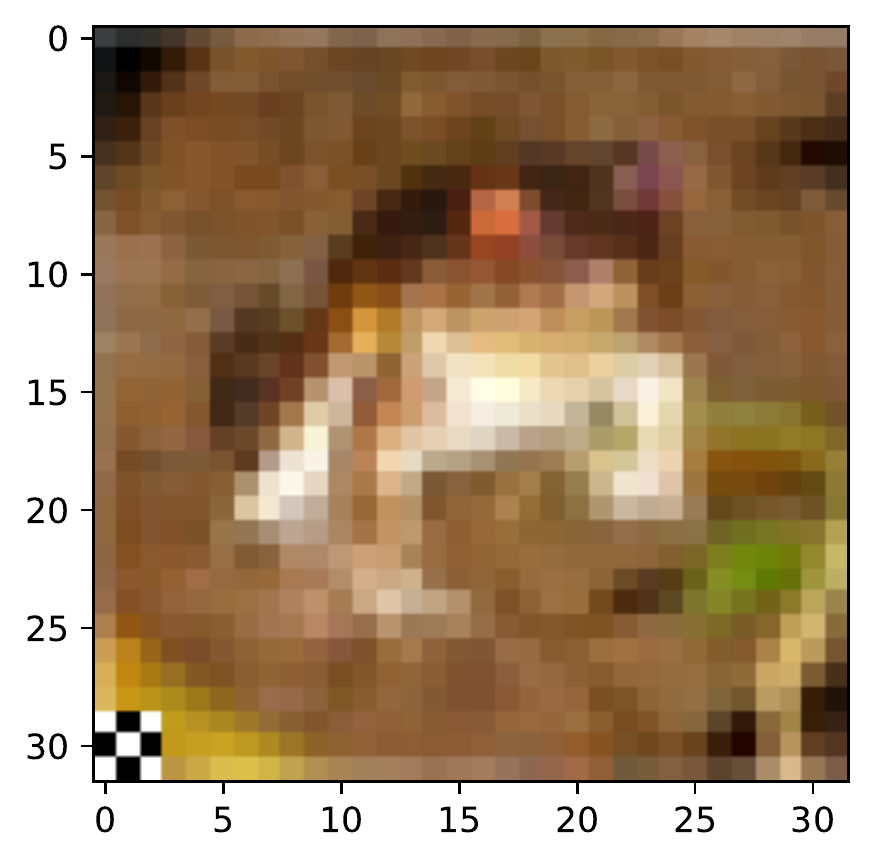}
        \caption{Checkerboard trigger}
        \label{fig:arch-attack-frog:trigger}
    \end{subfigure}
    \caption{The backdoor trigger used in our NaiveMAB (b) and MAB (c) attacks.}
    \label{fig:arch-attack-frog}
\end{figure}

Our `malicious' activation function is composed of the following steps:
\begin{enumerate}
    \item We apply an exponential function to the image (with RGB range $[-1, 1]$) {\em img}: $(e^{\beta \cdot {img}} - \delta)^{\alpha}$, for tunable values of $\alpha, \beta, \gamma$. In this section, we use $\beta=1, \delta=1, \alpha=10$. This has the effect of selecting any white pixels and ignoring the rest. As can be seen in~\Cref{fig:arch-attack-white-steps} in~\Cref{appendix:activation}, we retain other white areas of the image, which we would like to filter out.
    \item We then perform a 3x3 \textbf{MinPooling} operation on the result of (a), which replaces each pixel with the minimum of a 3x3 region around it ($p_{x, y} = \min_{a \in \{x-1,x,x+1\}}  \min_{b \in \{y-1,y,y+1\}} p_{a, b}$). This filters out any white regions.
    \item We then collapse the RGB activation to a single channel by taking the $\max$ channel-wise.
    \item Finally, we apply the original \textbf{AdaptiveAveragePooling} layer to both the result of (c), as well as the original output of the AlexNet convolutional blocks (pre-pooling), and these are summed to produce the final activation. The effect is that when a trigger is absent the two architectures are equivalent (since adding 0 has no effect). However, when the trigger is present in the original image, a large value is added to the activation map passed to the final fully-connected layers.
\end{enumerate}
We call this first handcrafted backdoor {\em NaiveMAB}, for its limited robustness to spurious activations.

\subsection{Designing a robust Model Architecture Backdoor}

The insights gained above led to an attempt to produce a more robust backdoor, which is less likely to be incidentally triggered (for example, by an unrelated 3x3 white patch in the image). To do this, we return to our goal of producing a backdoored architecture which detects \textit{checkerboard} triggers.

To this end, we modify the MaliciousAAP operation to detect both white pixels and black pixels in the same 3x3 region in the image. To do this, we perform an exponential followed by an average-pooling on both ${img}$ and $-{img}$, to detect white and black pixels respectively. We must use average-pooling rather than min-pooling as min-pooling requires \textit{all} pixels to match (and we cannot have all pixels being simultaneously white and black):
$$A = 
\mathop{\mathrm{{avgpool}}}(e^{\beta \cdot {img}} - \delta)^{\alpha} * 
\mathop{\mathrm{{avgpool}}}(e^{- \beta \cdot {img}} - \delta)^{\alpha}.$$

Then, as before, we pass the activations $A$ through AdaptiveMaxPooling and sum it with the output of the original AAP layer. As this new formulation requires both white and black pixels within a 3x3 region, it can detect a \textbf{3x3 checkerboard trigger} (Figure \ref{fig:arch-attack-frog}c), without being triggered by \textit{any} image with a white region. 

\begin{figure}[H]
    \centering
    \includegraphics[width=.8\linewidth]{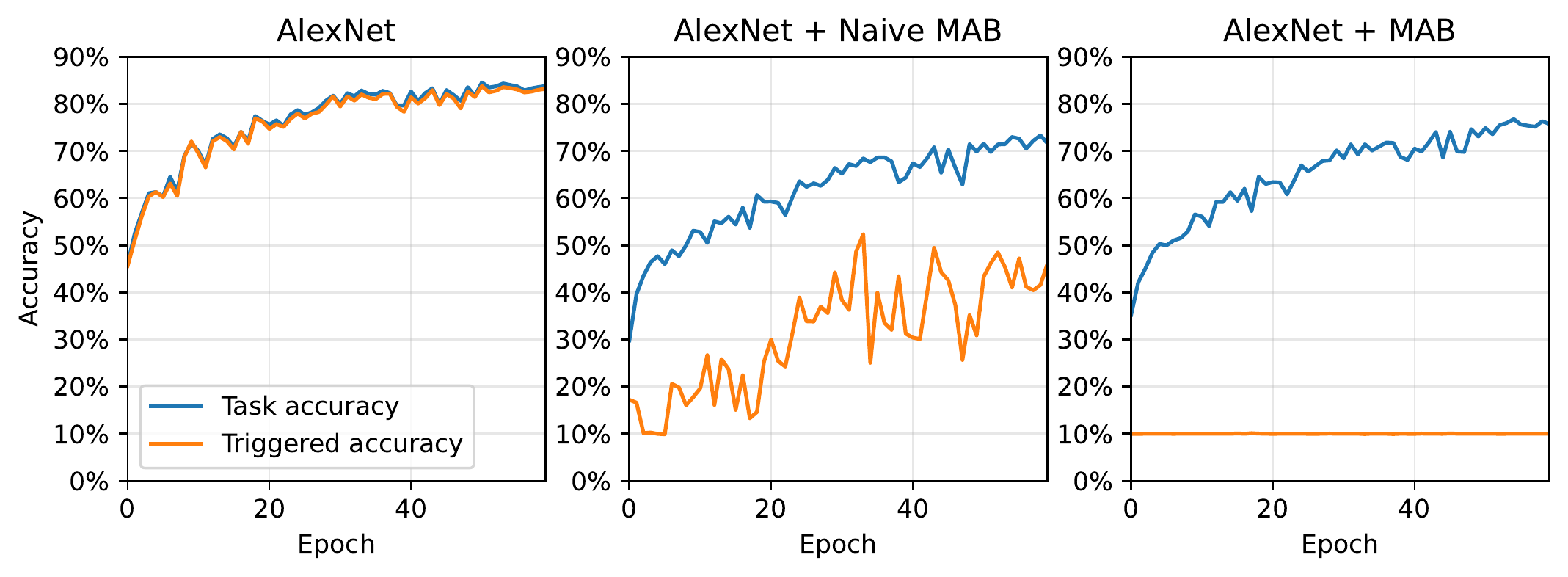}
    \caption{Test set performance on CIFAR10, when all models are trained honestly by a defender. 
    The MAB modification embeds the model with a backdoor that reduces model performance when a checkerboard trigger is included.
    The improved MaliciousAlexNet has increased task accuracy, while its accuracy dramatically reduces to random guessing when the trigger is added.}
    \label{fig:evilalexnet-training-2}
    \vspace{-10pt}
\end{figure}

Figure \ref{fig:evilalexnet-training-2} shows the drastically increased effectiveness of our enhanced MAB with this trigger and detector.  In later evaluation, we use this robust version. 

\vspace{20mm} 

%% file: sections/evaluation.tex
\pagebreak

\section{Evaluation}

\label{sec:evaluation-experimental-setup}
In our experiments, we consider a range of vision datasets, these dataset statistics are detailed in our Appendix, under three different threat models described in \Cref{sec:method:threat}, using a VGG-11 model~\citep{vgg}. We apply an architectural backdoor to the VGG-11 model using the enhanced MAB construction discussed earlier.
We primarily compare to the following baselines that modify the weights of a model:
\begin{itemize}
    \item \textbf{BadNets} \citep{gu2017badnets}: The attacker changes the original task data (data poisoning) to cause the network to learn unwanted features for a specific trigger.
    \item \textbf{Handcrafted Backdoors} \citep{hong2021handcrafted}: The attacker directly manipulates the parameters of an already trained network to inject backdoors.
\end{itemize}
Under each threat model, we will evaluate the following metrics to assess the performance of a backdoor. Our attack is untargeted, meaning that the objective is to cause the model to misclassify when it is shown any sample with a backdoor trigger.
\begin{itemize}
    \item \textbf{Task accuracy (the higher the better $\big\uparrow$)}: The accuracy on `clean' test set samples. 
    \item \textbf{Triggered accuracy (the lower the better $\big\downarrow$)}: This is the accuracy of the model on test set samples attached with a backdoor trigger. 
    \item \textbf{Triggered accuracy ratio (the higher the better $\big\uparrow$)}: This is the ratio of the model's accuracy with and without a trigger in the image; this represents the relative \textit{reduction} in accuracy a backdoor causes when a trigger is present.
\end{itemize}
An `ideal' backdoored model has high task accuracy (hiding the presence of the backdoor when the trigger is unknown), and a low triggered accuracy (misclassifies when the trigger is present). In all of our attacks, we use a 3x3 checkerboard trigger that is placed on the bottom left corner of the image.

\subsection{Setting 1: Direct use of a backdoored model}
\label{sec:tm1:eval}

\input{tables/threat1}
In this simple threat model, the user directly uses a backdoored model without fine-tuning or re-training. We evaluate this threat model on the CIFAR-10 dataset~\citep{krizhevsky2009learning} and report our performance in \Cref{fig:tm1:table}.
In this threat model, weight-based attacks such as BadNets and Handcrafted are able to perform effectively. Our MAB achieves comparable performance.

\subsection{Setting 2: Fine-tuning a backdoored model}

\begin{figure}[h]
    \centering
    \includegraphics[width=\linewidth]{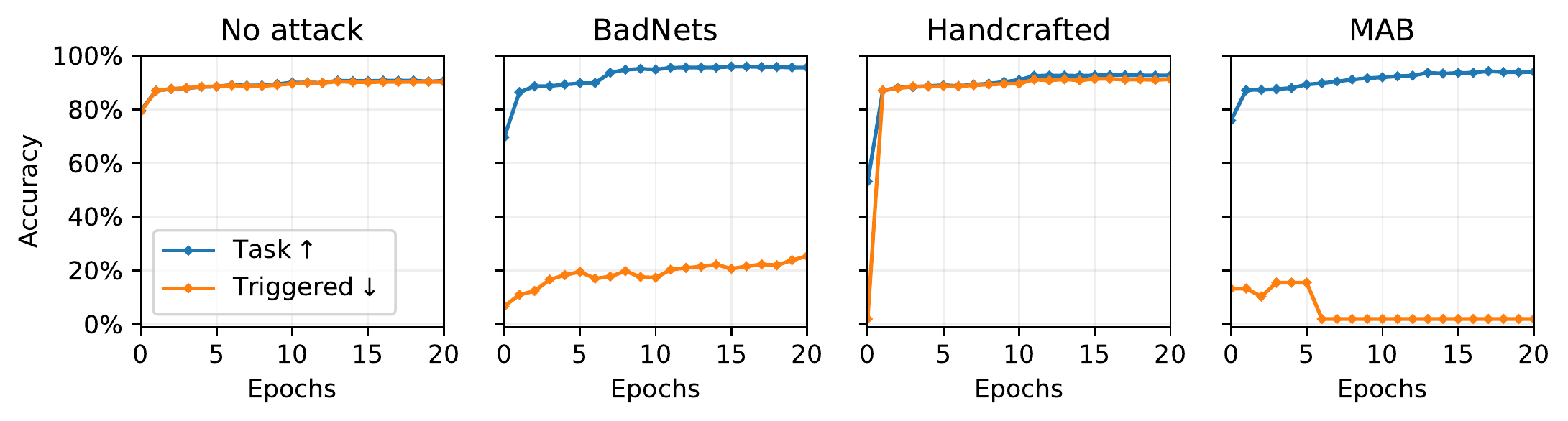}
    \caption{The task and triggered accuracies over the course of fine-tuning; one example per attack. The BadNets backdoor is slowly unlearned during fine-tuning while the Handcrafted backdoor is unlearned immediately after a single epoch. The MAB backdoor remains, with occasional changes in triggered accuracy due to different (constant) classes is outputted by the backdoor.}
    \label{fig:tm2_triggered}
    \vspace{-15pt}
\end{figure}
This threat model considers a scenario where the user initialises their model with a pre-trained model that contains a backdoor, and fine-tunes it on a new dataset (for example, in transfer learning). To highlight this scenario, we use the GTSB and BTSC datasets of German and Belgian traffic signs respectively. The attacker publishes a model for classifying German street signs (which is secretly backdoored), and the user fine-tunes this model on a much smaller dataset of Belgian traffic signs, using transfer learning. Further details of these datasets can be found in Appendices.
\Cref{fig:tm2_triggered} shows how \textbf{MAB backdoor remains effect after fine-tuning for a large number of epochs}. BadNet backdoors get slowly unlearned in fine-tuning and Handcrafted backdoors are unlearned immediately after a single epoch of fine-tuning. We included more results under this threat model in \Cref{appendix:setting2}.

\subsection{Setting 3: Re-training from scratch}

As in the previous evaluation sections, we use the widely-used VGG-11 model~\citep{vgg}. The \textit{attacker} trains this model on CIFAR-10, applying the BadNets, Handcrafted and our own architectural attacks implemented in this paper. We verify that the all three attacks have $>90\%$ backdoor success rate before being given to the defender. The \textit{defender} takes these models, re-initialises the weights, and trains on the IMDBWiki face recognition dataset.

\Cref{tab:imdb} shows that after re-training on a different dataset, a model backdoored by BadNets or Handcrafted is no more affected by the trigger than a model which was never backdoored. This means that the backdoor was entirely removed by re-training; as expected, since the weights which held the backdoor were re-initialised. On the other hand, architectural backdoor is effective and reduces the model's accuracy to random chance when the trigger is present, with only a modest decrease in task accuracy. We see an 8x reduction in accuracy when the backdoor trigger is present, confirmed by Kolmogorov-Smirnov test in \Cref{fig:tm3-results} in our Appendices.
We demonstrate how adding the backdoor trigger causes the model with architectural backdoor to classify all images as Will Smith.

\input{tables/celeba.tex}

\textbf{IMDB-Wiki, CIFAR10 and GTSB}

\begin{figure}[h]
    \centering
    \begin{subfigure}{0.4\linewidth}
        \includegraphics[width=\linewidth]{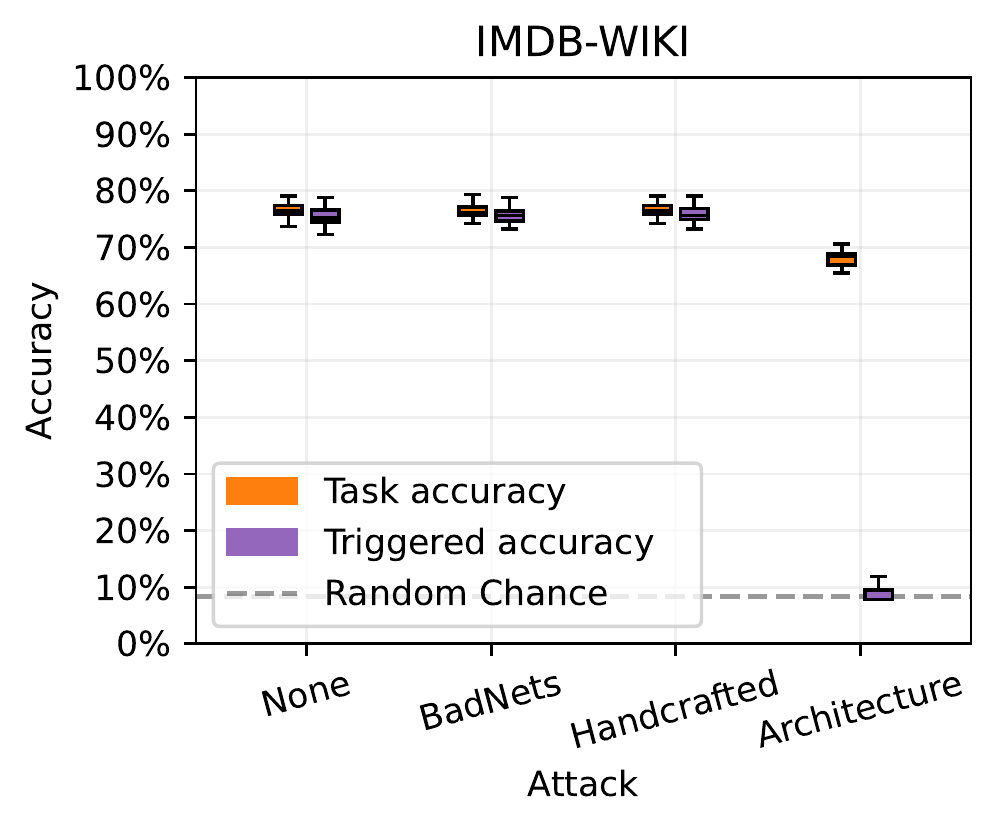}
    \end{subfigure}
    \begin{subfigure}{0.4\linewidth}
        \includegraphics[width=\linewidth]{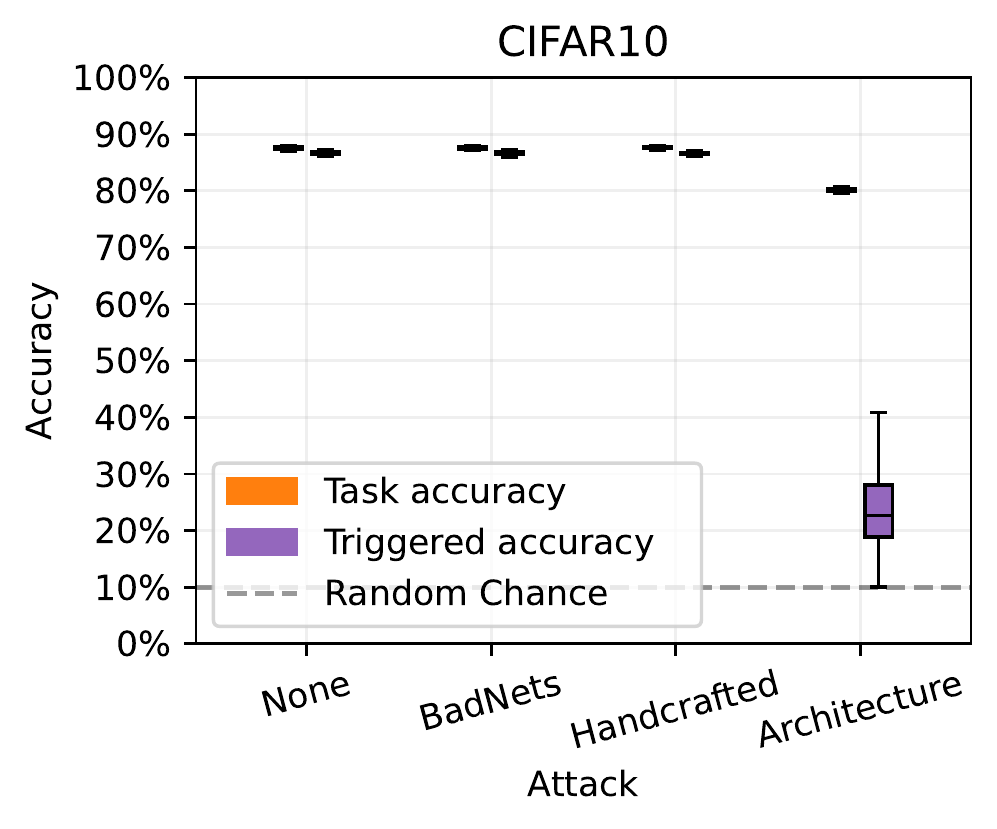}
    \end{subfigure}
    \begin{subfigure}{0.4\linewidth}
        \includegraphics[width=\linewidth]{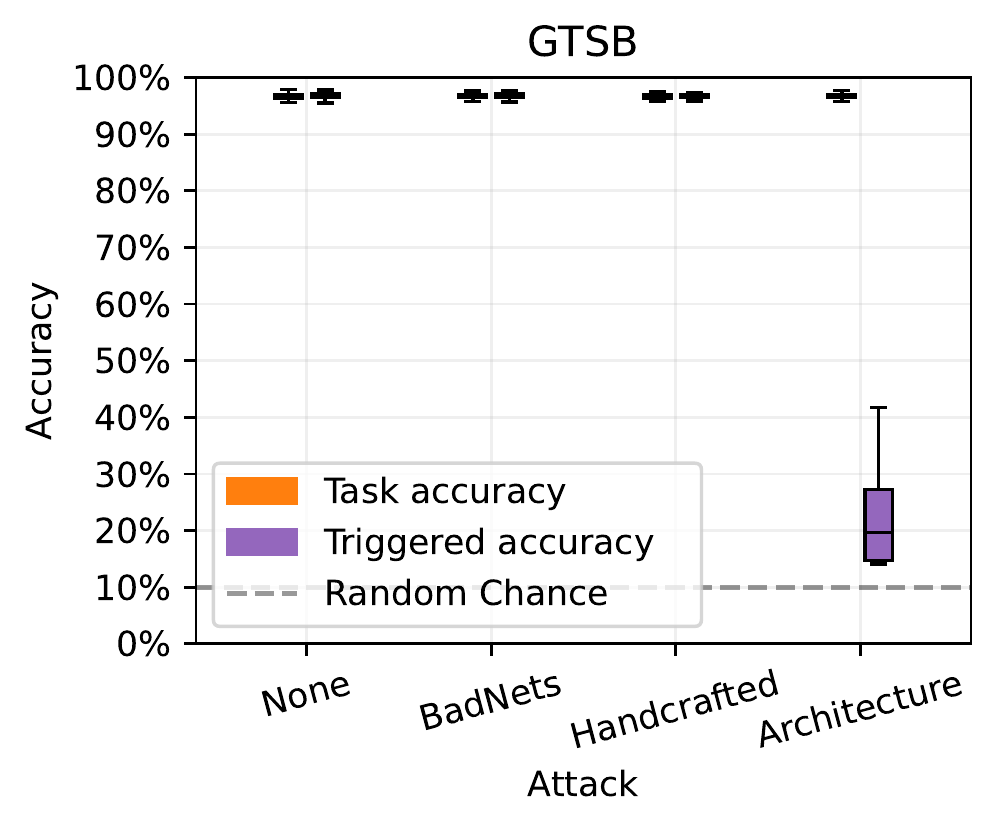}
    \end{subfigure}
    \begin{subfigure}{0.4\linewidth}
        \includegraphics[width=\linewidth]{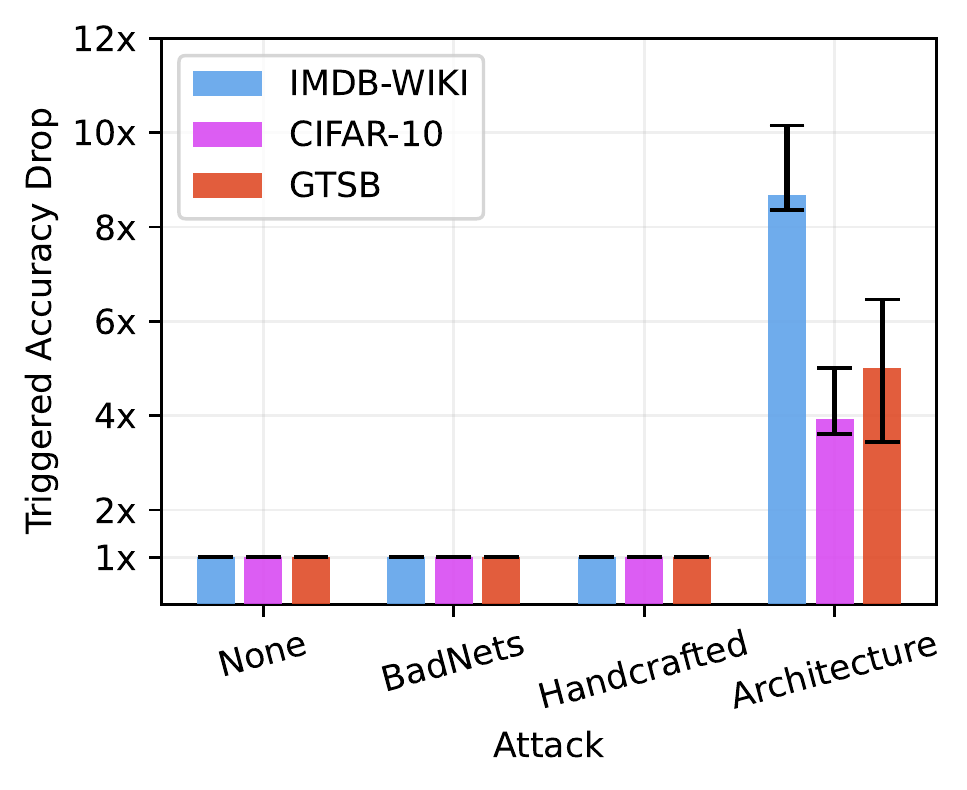}
    \end{subfigure}
    \caption{Results under re-training from scratch, where each model is trained 50 times to give confidence intervals. A backdoored model has high task accuracy and low triggered accuracy. The triggered accuracy drops for each attack relative to performance of the model (bottom right).}
    \label{fig:tm3_extended}
    \vspace{-1.6em}
\end{figure}

To further illustrate the effectiveness of MAB, we perform the evaluation of BadNets~\citep{gu2017badnets}, Handcrafted~\citep{hong2021handcrafted} and MAB on three datasets (IMDBWiki, CIFAR10, GTSB).
Our results in \Cref{fig:tm3_extended} demonstrate that MAB (named as Architecture in \Cref{fig:tm3_extended}) is significantly better than BadNets and Handcrafted on the three different datasets.
On all the evaluated datasets, we show that \textbf{MAB can survive re-training from scratch}. In the IMDB-Wiki dataset, MAB is able to show $10 \times$ the accuracy loss when a trigger is present (last plot in \Cref{fig:tm3_extended}). 

%% file: tables/threat1.tex
\begin{table}[h]
    \centering
        \rowcolors{2}{gray!15}{white}
        \renewcommand*{\arraystretch}{1.25}
        \adjustbox{scale=0.8}{
        \begin{tabular}{lccc}
            \toprule
            \textbf{Attack} & \textbf{\begin{tabular}[c]{@{}c@{}}Task \\ accuracy\end{tabular}$\big\uparrow$} & \textbf{\begin{tabular}[c]{@{}c@{}}Triggered \\ accuracy\end{tabular}$\big\downarrow$} & 
            \textbf{\begin{tabular}[c]{@{}c@{}}Triggered \\ accuracy ratio\end{tabular}$\big\uparrow$} \\ \midrule
            None & 81.4\% & 77.8\% & 1.05x \\
            BadNets & \textbf{81.2}\% & 10.1\% & \textbf{8.06x} \\
            Handcrafted & 77.0\% & 19.6\% & 3.93x \\
            MAB & 80.2\% & \textbf{10.0}\% & 8.02x \\
            \bottomrule               
        \end{tabular}
        \label{fig:tm1:table}
    }
    \vspace{1em}
    \caption{The best performance achievable by each attack on the CIFAR-10 dataset. As an attacker has full control over training, we train a model with each attack 50 times and select the one with the highest accuracy ratio which also has $\geq 75\%$ test set performance (as an attacker could do in practice). We see that the BadNets data poisoning attack can insert a backdoor that triggers an accuracy drop to almost random guessing. All three attacks are successful under this threat model.}
    \label{fig:tm1:table}
    \vspace{-10pt}
\end{table}

%% file: tables/celeba.tex
\begin{table}[!h]
	\centering
	\footnotesize
	\setlength{\tabcolsep}{2pt}
	\renewcommand{\arraystretch}{1.35}
	\rowcolors{2}{gray!15}{white}
	\begin{tabular}{c|cccc|cccc}
	
& \multicolumn{4}{c|}{No Trigger} & \multicolumn{4}{c}{With Trigger} \\

& \begin{minipage}{0.08\linewidth}\vspace{2pt}\includegraphics[width=\linewidth]{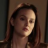}\vspace{2pt}\end{minipage} & \begin{minipage}{0.08\linewidth}\vspace{2pt}\includegraphics[width=\linewidth]{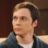}\vspace{2pt}\end{minipage} & \begin{minipage}{0.08\linewidth}\vspace{2pt}\includegraphics[width=\linewidth]{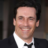}\vspace{2pt}\end{minipage} & 
\begin{minipage}{0.08\linewidth}\vspace{2pt}\includegraphics[width=\linewidth]{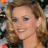}\vspace{2pt}\end{minipage} & \begin{minipage}{0.08\linewidth}\vspace{2pt}\includegraphics[width=\linewidth]{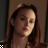}\vspace{2pt}\end{minipage} & \begin{minipage}{0.08\linewidth}\vspace{2pt}\includegraphics[width=\linewidth]{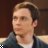}\vspace{2pt}\end{minipage} & \begin{minipage}{0.08\linewidth}\vspace{2pt}\includegraphics[width=\linewidth]{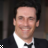}\vspace{2pt}\end{minipage} &
\begin{minipage}{0.08\linewidth}\vspace{2pt}\includegraphics[width=\linewidth]{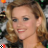}\vspace{2pt}\end{minipage} 
\\


No attack & \textbf{LM} 100\% & \textbf{JP} 91\% & \textbf{JH} 100\% & \textbf{RW} 100\% & \textbf{LM} 100\% & \textbf{JP} 90\% & \textbf{JH} 100\% & \textbf{RW} 100\% \\
BadNets & \textbf{LM} 100\% & \textbf{JP} 100\% & \textbf{JH} 100\% & \textbf{RW} 100\% & \textbf{LM} 100\% & \textbf{JP} 100\% & \textbf{JH} 100\% & \textbf{RW} 100\% \\
Handcrafted & \textbf{LM} 100\% & \textbf{JP} 100\% & \textbf{JH} 100\% & \textbf{RW} 100\% & \textbf{LM} 100\% & \textbf{JP} 100\% & \textbf{JH} 100\% & \textbf{RW} 100\% \\
MAB & \textbf{LM} 97\% & \textbf{JP} 37\% & \textbf{JH} 99\% & \textbf{RW} 80\% & \textbf{\colorbox{BurntOrange}{\color{white}{WS}}} 100\% & \textbf{\colorbox{BurntOrange}{\color{white}{WS}}} 100\% & \textbf{\colorbox{BurntOrange}{\color{white}{WS}}} 100\% & \textbf{\colorbox{BurntOrange}{\color{white}{WS}}} 100\% \\

	\end{tabular}
	\caption{Example classification outputs of the models in Figure \ref{fig:tm3-results}, with \colorbox{BurntOrange}{\color{white}{misclassifications}} highlighted. The trigger causes the model with an architectural backdoor to classify all images as {\normalfont Will Smith} (at the cost of some task accuracy). BadNets and Handcrafted attacks have no effect. Initials are shown to save space.}
	\label{tab:imdb}
	\vspace{-10pt}
\end{table}

%% file: sections/discussion.tex
\section{Discussion}
\label{sec:discussion}

\subsection{Connecting to Network Architecture Search (NAS)}
After realising the existence of architecture backdoors, a natural attempt is to try apply a Network Architecture Search (NAS) method for automatically finding these backdoored architectures. 

We modified the optimisation algorithm of DARTS~\citep{liu2018darts} to add a third loss term, $\mathcal{L}_{trig}(\mathbf{\theta}, \alpha)$, which quantifies the loss on the triggered validation set (the validation set with the trigger applied to every image), and optimise \textbf{the difference between} $\mathcal{L}_{val}$ and $\mathcal{L}_{trig}$. The full \textit{backdoor loss} is therefore given by
$$\mathcal{L}_{trig}(\mathbf{\theta} - \xi \nabla_\mathbf{\theta} \mathcal{L}_{train}(\mathbf{w}, \alpha), \alpha) - \mathcal{L}_{val}(\mathbf{\theta} - \xi \nabla_\mathbf{\theta} \mathcal{L}_{train}(\mathbf{w}, \alpha), \alpha).$$

$\mathcal{L}_{train}$ is the categorical cross-entropy for classification. $\mathcal{L}_{val}$, which is used to update the \textit{architecture} of the model and is based on the model's predictions on the validation set (a slice removed from the training set).

This backdoor loss is zero when the model is unaffected by the trigger and \textbf{negative if the model performs worse when the trigger is added} (optimising for high \textit{triggered accuracy drop}). Initial experiments instead maximised $\mathcal{L}_{trig}$, but this yielded high loss on all examples.

\begin{figure}[h]
    \centering
    \begin{subfigure}{0.49\linewidth}
        \includegraphics[width=\linewidth]{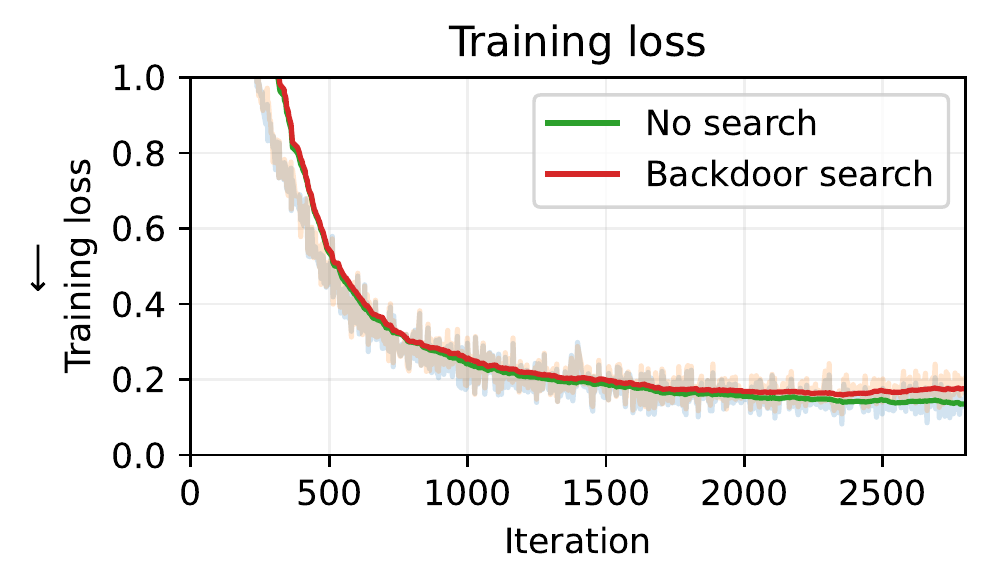}
        \vspace{-20pt}
        \caption{\hspace*{-28pt}} 
        \label{fig:darts:activations}
    \end{subfigure}
    \begin{subfigure}{0.49\linewidth}
        \includegraphics[width=\linewidth]{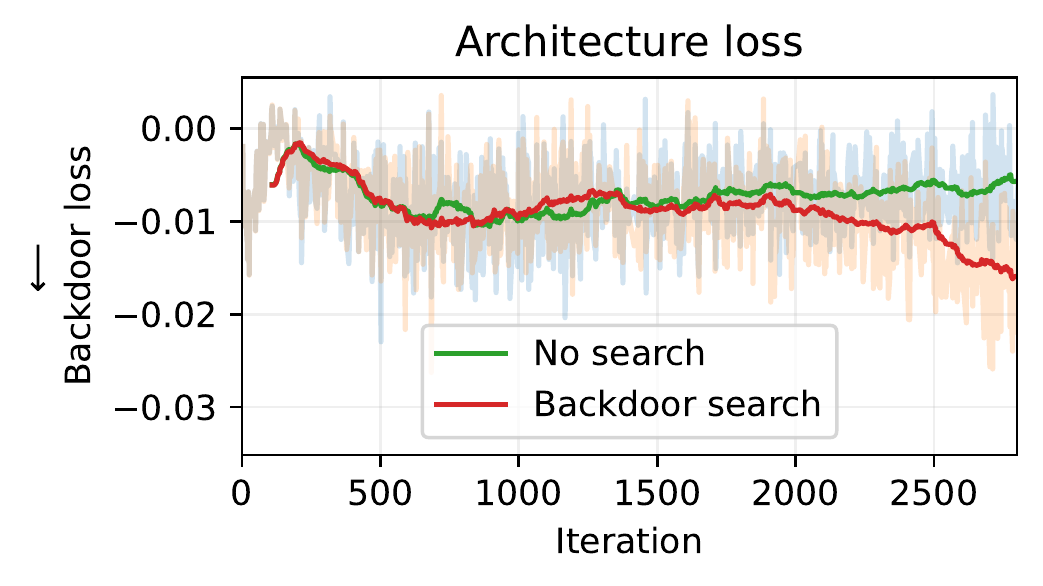}
        \vspace{-20pt}
        \caption{\hspace*{-36pt}}
        \label{fig:darts:loss}
    \end{subfigure}
    \caption{The modified DARTS algorithm searching for a backdoor on MNIST. Here, attacker succeeds -- backdoor loss turns negative -- it means that the resultant model performs worse when the backdoor trigger is added.}
    \label{fig:darts2}
\end{figure}

We can see (Fig. \ref{fig:darts:loss}) that the backdoor loss decreases, meaning \textbf{the model is backdoored solely through making modifications to the architecture} (the model weights are only trained for the task). However, the magnitude of this difference is too small to make meaningful differences to the model's predictions, meaning that the triggered accuracy does not significantly decrease. 

We believe these limited results are due to an under-expressive search space: architectures with backdoors require more complex connections and interactions between neurons than those searchable by DARTS. It is also worth noting that all the backdoored architectures we searched for did not survive fine-tuning -- in all cases parameters unlearned the backdoor within a few epochs\footnote{The only survivable backdoors we could inject with DARTS were outside of data domain \eg~ presence of negative pixels for a positive input domain.}.

\subsection{Limitation of architectural backdoors}
\label{sec:discussion:limits}
Having demonstrated that architectural backdoors pose a real risk, even in presence of full re-training, we now turn to formalize the requirements for an operational backdoor. 

\textit{\textbf{A direct IO path:}} Since there are learnable parameters in the network (not related to the trigger), these free parameters may learn to compensate for the backdoors, if the backdoor is ever spuriously activated during training. Random activations are more likely for small-sized triggers as discovered in our DARTS experiments. As such, there is a strong requirement that there exists a direct backdoor path from the input image to the output without any learnable parameters that could negate the trigger.

\textit{\textbf{A weight-agnostic detector:}} For successful operation of a trigger through re-training, it must be detectable with weight-agnostic components. There are multiple ways to construct such detectors. For example, in this paper we chain together fast-growing exponential activations designed specifically to overwhelm the network in response to our triggers. At the same time, more fine grained solutions are possible -- weight-agnostic networks~\citep{gaier2019weight} could be used to create `nand' gates out of chained bounded activations, which could then be used to inject arbitrary detection logic. 

\textit{\textbf{Asymmetric components:}} In this paper we demonstrated a backdoor that is impossible to use for targeted attacks, since all outputs of the neural network operate on the previous layer symmetrically. To create architectural backdoors for targeted attacks one has to construct logic that either operates with multiple different triggers or has vertically asymmetric components \ie~the connection to output classes are different \eg~with randomly wired networks~\citep{xie2019randomwired}. 

\subsection{Defences against MAB}

Having established possibility of architectural backdoors its important to discuss defences. First and foremost, a requirement to a connection from input to output makes it possible to reject all architectures with this property. Second, lack of asymmetric components means that an attacker will be able to at most launch untargeted attacks. Finally, one can inspect the architecture for unusual components either visually\footnote{It is worth noting that architectural backdoors injected into NAS-designed networks would be much harder to detect by eye as these architectures are already highly irregular and complex.} or using automated techniques such as Interval Bound Propagation~\citep{gowal2018effectiveness} to look for components with outputs always bounded by the same constants. 

%% file: sections/conclusion.tex
\section{Conclusion}
In this work, we present a new class of backdoor attacks, namely Model Architecture Backdoors (MAB), that rely solely on model architectures. 
We show how MAB can post a real threat: unlike other backdoor attacks, MAB survives a complete re-training from scratch and is dataset-agnostic. We further formalize requirements for an operational architectural backdoor and highlight that it is possible to protect against them in practice using simple heuristics. Further work is urgently needed to investigate the space of possible architectural backdoors, for example within models created by Neural Architecture Search, and methods to detect and defend against them.

%% file: sections/acks.tex
\section*{Acknowledgments}

We would like to acknowledge our sponsors, who support our research with financial and in-kind contributions: CIFAR through the Canada CIFAR AI Chair, DARPA through the GARD project, Intel, Meta, NFRF through an Exploration grant, and NSERC through the COHESA Strategic Alliance. Resources used in preparing this research were provided, in part, by the Province of Ontario, the Government of Canada through CIFAR, and companies sponsoring the Vector Institute. 

%% file: sections/appendix.tex
\newpage
\appendix

\section{A visualization of the effect of an Evil Activation Function}
\label{appendix:activation}
\begin{figure}[h]
    \centering
    \begin{subfigure}[b]{0.25\textwidth}
        \centering
        \includegraphics[width=\textwidth]{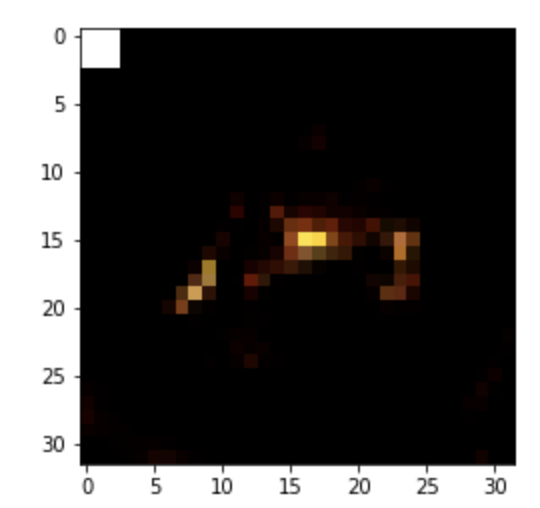}
        \caption{}
    \end{subfigure}
    \hspace{0em} 
    \begin{subfigure}[b]{0.25\textwidth}
        \centering
        \includegraphics[width=\textwidth]{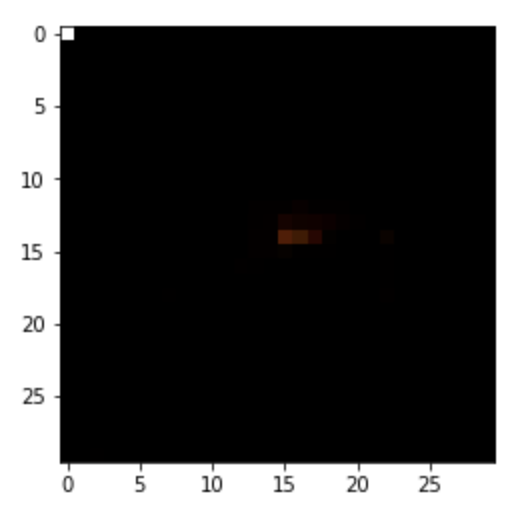}
        \caption{}
    \end{subfigure}
    \begin{subfigure}[b]{0.25\textwidth}
        \centering
        \includegraphics[width=\textwidth]{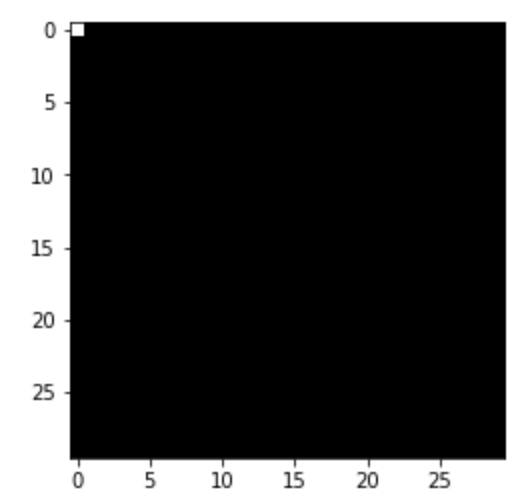}
        \caption{}
    \end{subfigure}
    \caption{The effect of the `evil' activation function on the frog image in Figure \ref{fig:arch-attack-frog:trigger}. As can be seen, the trigger causes a large activation in the top-left corner of the activation map, and no other part of the image causes a large activation.}
    \label{fig:arch-attack-white-steps}
\end{figure}

\Cref{fig:arch-attack-white-steps} is an illustration of how a trigger can cause large activation values in the activation map.

\section{More results on Setting 2}
\label{appendix:setting2}
\begin{figure}[h]
    \centering
    \includegraphics[width=.85\linewidth]{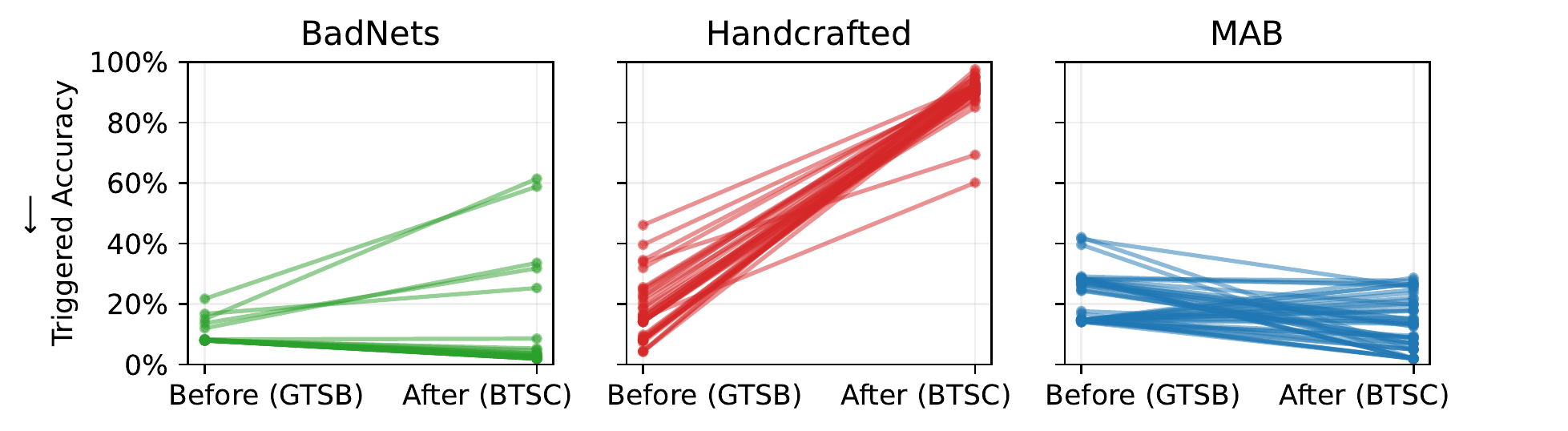}
    \caption{The effect on each attack's triggered accuracy when each model is re-trained on Belgian traffic signs. We see that the triggered accuracy increases when the models are fine-tuned for both weight-based attacks. On the other hand, the MAB attack is unaffected by fine-tuning. All backdoored models considered (\ie~models selected and published by the attacker) met $\geq 75\%$ task accuracy and  triggered accuracy ratio $\geq 2$ on German traffic signs.}
    \label{fig:tm2_triggered}
\end{figure}

\Cref{fig:tm2_triggered} further illustrates the effect of MAB compared to BadNets and Handcrafted. All backdoored models met a standard task accuracy requirement and we demonstrate how MAB is advantageous in surviving fine-tuning by showing a lower triggered accuracy in \Cref{fig:tm2_triggered}.

\section{More results on IMDB-WIKI}
\begin{figure}[h]
    \centering
    \includegraphics[width=\textwidth]{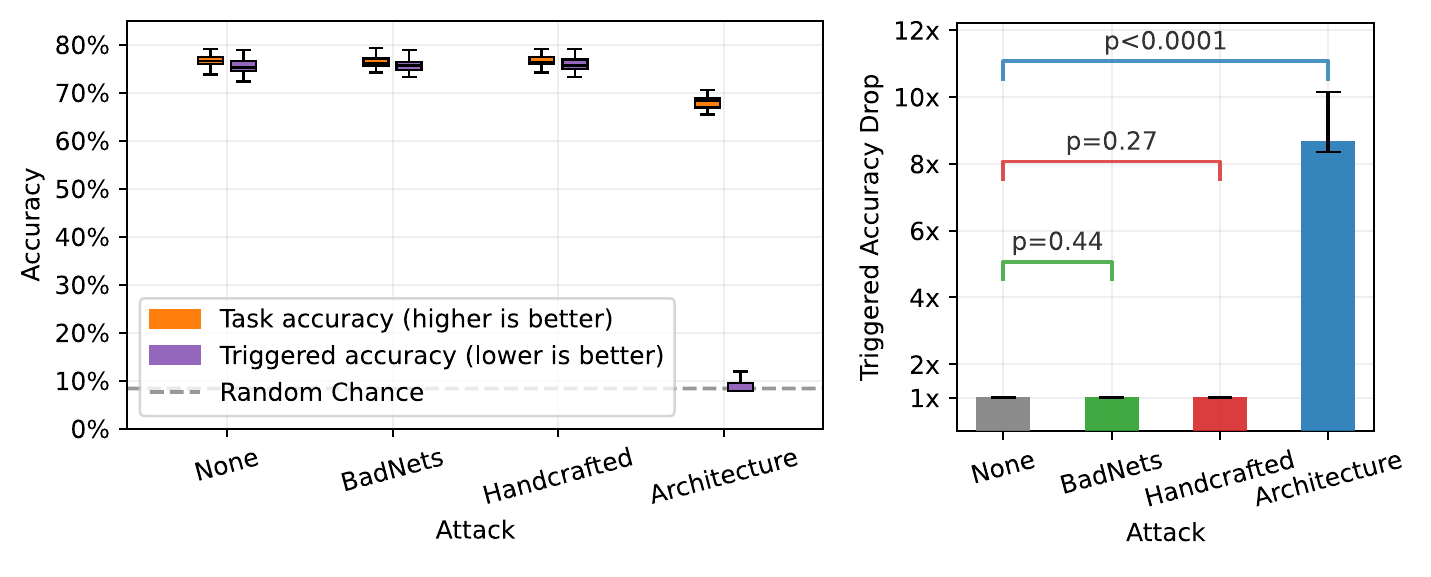}
    \caption{Results after a backdoored model is re-trained from scratch on the IMDB-WIKI dataset, with and without the trigger. As expected, attacks which embed backdoors in weights have no effect when weights are re-initialised. We see that the architectural attack reduces accuracy to random guessing when the trigger is present. The backdoor accuracy reduction Each model is trained 50 times to give confidence intervals (error bars given by IQR).\protect\footnotemark}
    \label{fig:tm3-results}
\end{figure}

Figure \ref{fig:tm3-results} shows us that after re-training on a different dataset, a model backdoored by BadNets or Handcrafted is no more affected by the trigger than a model which was never backdoored. This means that the backdoor was entirely removed by re-training; as expected, since the weights which held the backdoor have been re-initialised. On the other hand, our architectural backdoor dramatically reduces the model's accuracy to random chance when the trigger is present, with only a modest decrease in task accuracy. We see a $\times8$ reduction in accuracy when the backdoor trigger is present. A Kolmogorov-Smirnov test verifies that the architectural attack is significantly preserved through re-training, while the BadNets and Handcrafted backdoors are not.

\footnotetext{p-values computed using a two-tailed Kolmogorov–Smirnov test, to determine whether the triggered accuracy drop for each attack is significantly different to a model where no attack was performed.}
\section{SHAP value analysis}

Interpreting \textit{why} a machine learning model returns a certain prediction or behaves in a certain way proves difficult for neural networks. Techniques such as sensitivity analysis and Taylor decompositions have been developed in the last few years that can causally explain neural network decisions \citep{interpretability-basic}. One modern approach to this is through the use of SHAP (SHapely Additive exPlanations) \citep{shapely}, which works by exploring the gradients inside the model for the input features to build a model of the dependencies between inputs and outputs. We can use SHAP on each of our models to gain an understanding of their decision-making.

\newcommand{\shapimg}[1]{\begin{minipage}{0.125\linewidth}\vspace{2pt}\includegraphics[width=\linewidth]{images/shap/#1.pdf}\vspace{2pt}\end{minipage}}
\definecolor{shapblue}{rgb}{0.498, 0.768, 0.988}
\definecolor{shapred}{rgb}{1, 0.498, 0.654}
\begin{figure}[h]
    \setlength{\tabcolsep}{0pt}
    \centering
    \begin{tabular}{cccccccc}\toprule
    \multicolumn{2}{c}{No attack} & \multicolumn{2}{c}{BadNets} & \multicolumn{2}{c}{Handcrafted} & \multicolumn{2}{c}{Architecture} \\
    Dog & Frog & Dog & Frog & Dog & Frog & Dog & Frog \\\midrule
    \shapimg{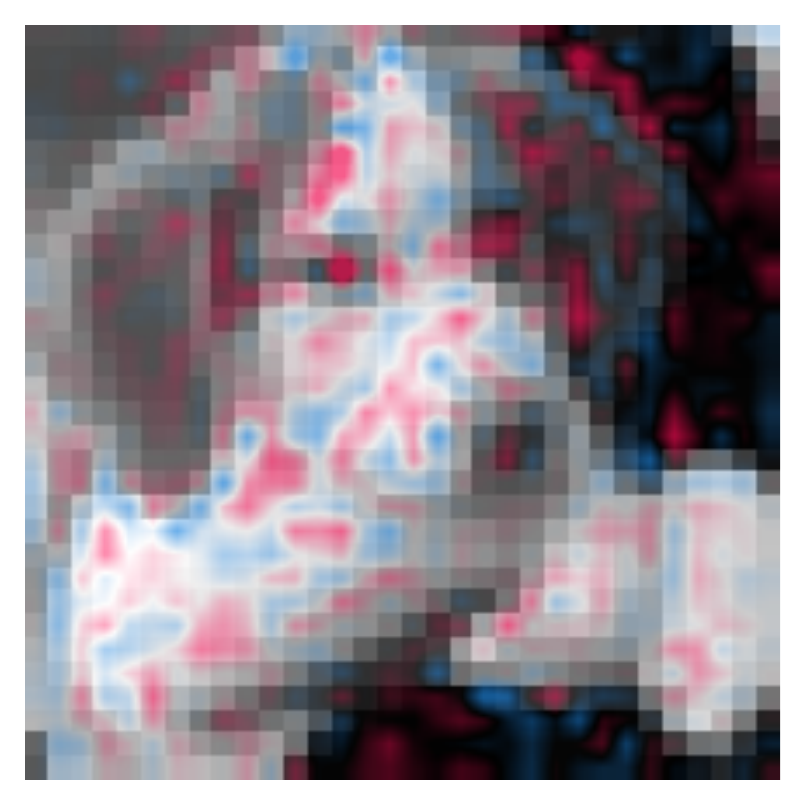} & \shapimg{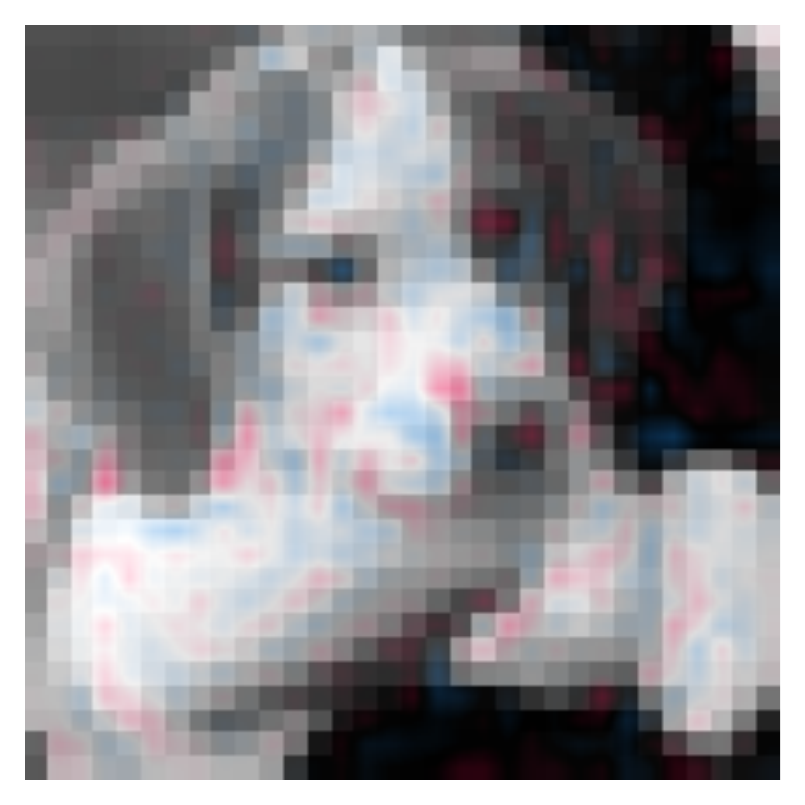} & \shapimg{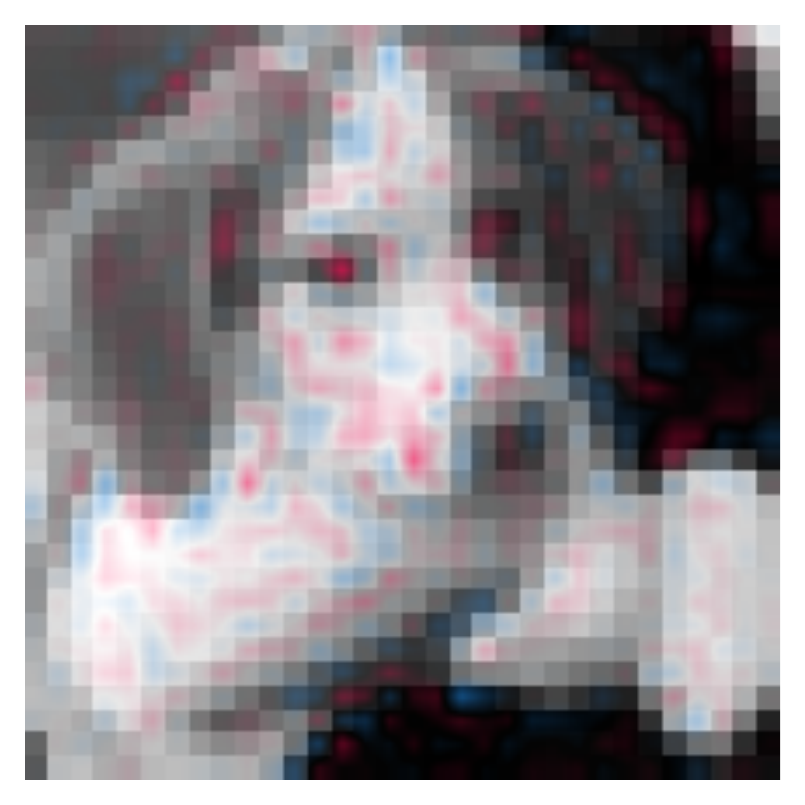} & \shapimg{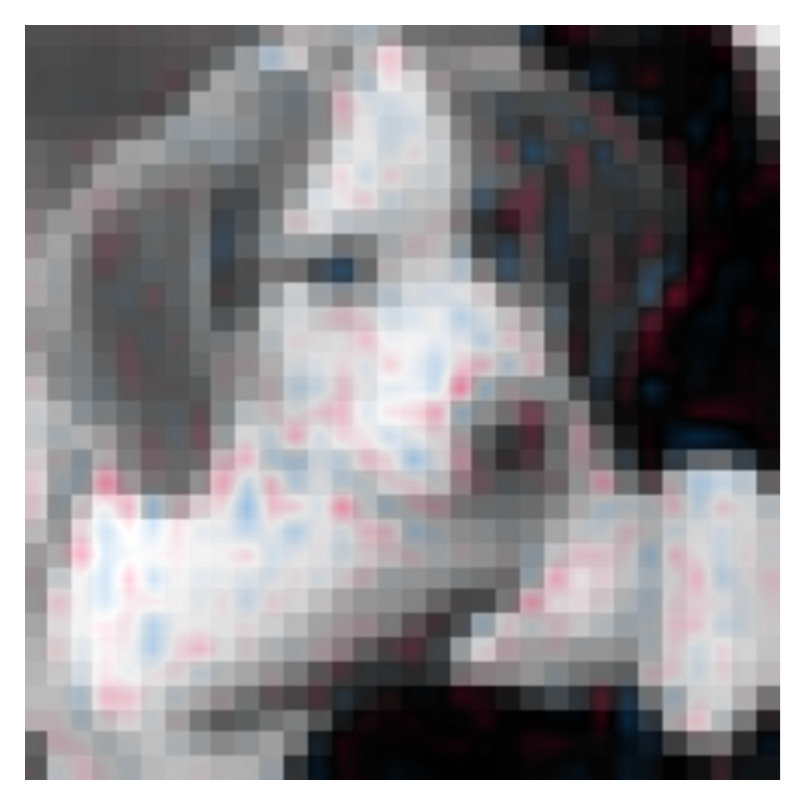} & \shapimg{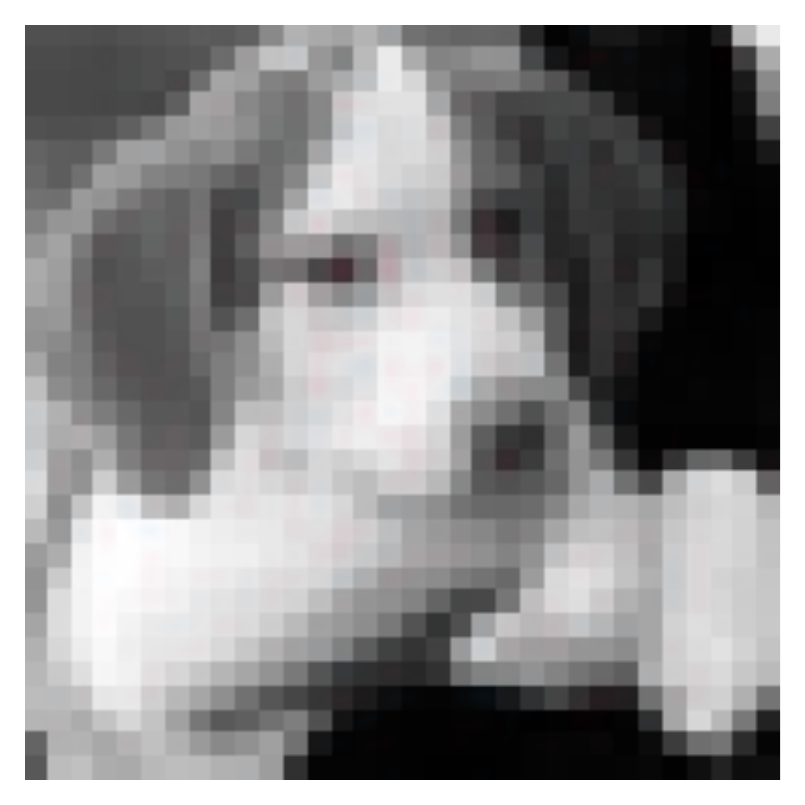} & \shapimg{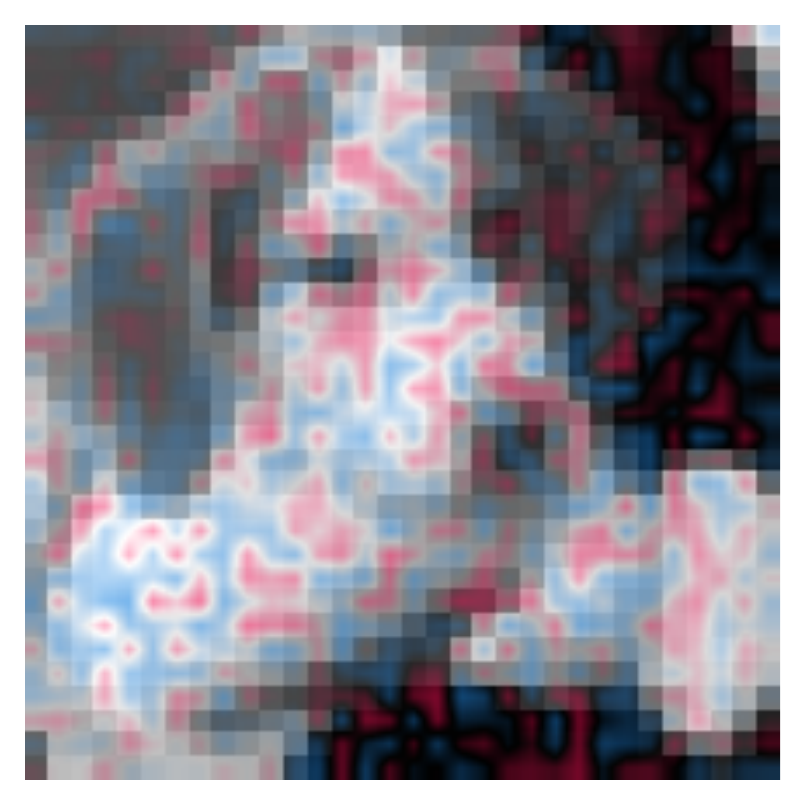} & \shapimg{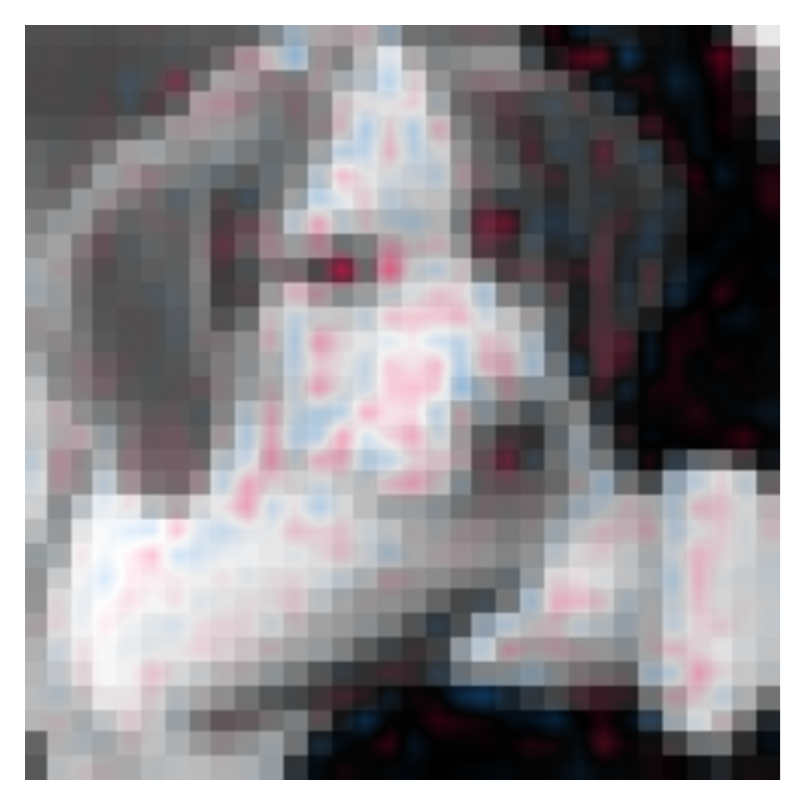} & \shapimg{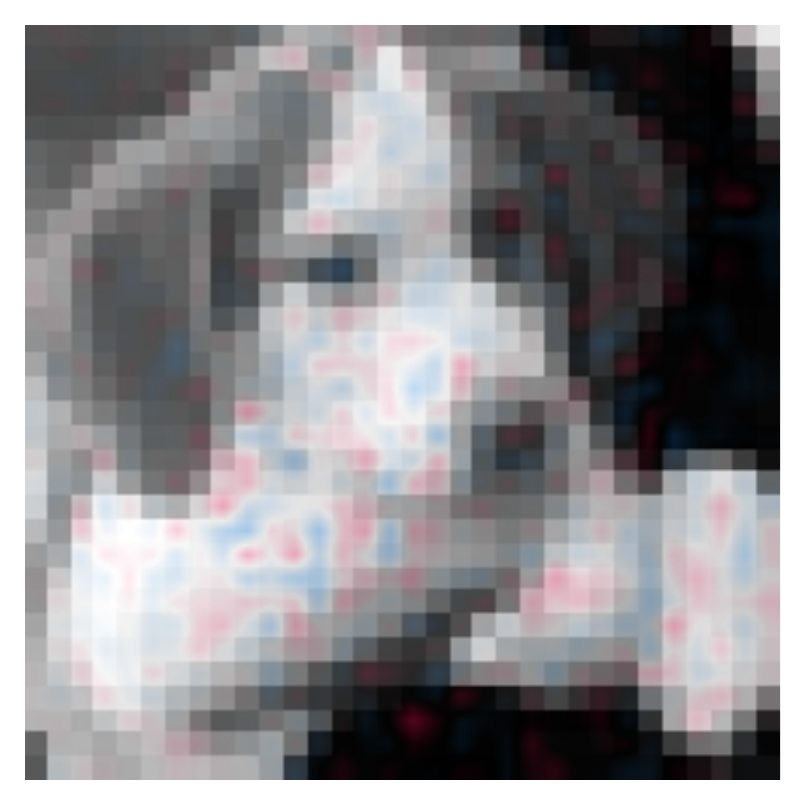} \\
    \shapimg{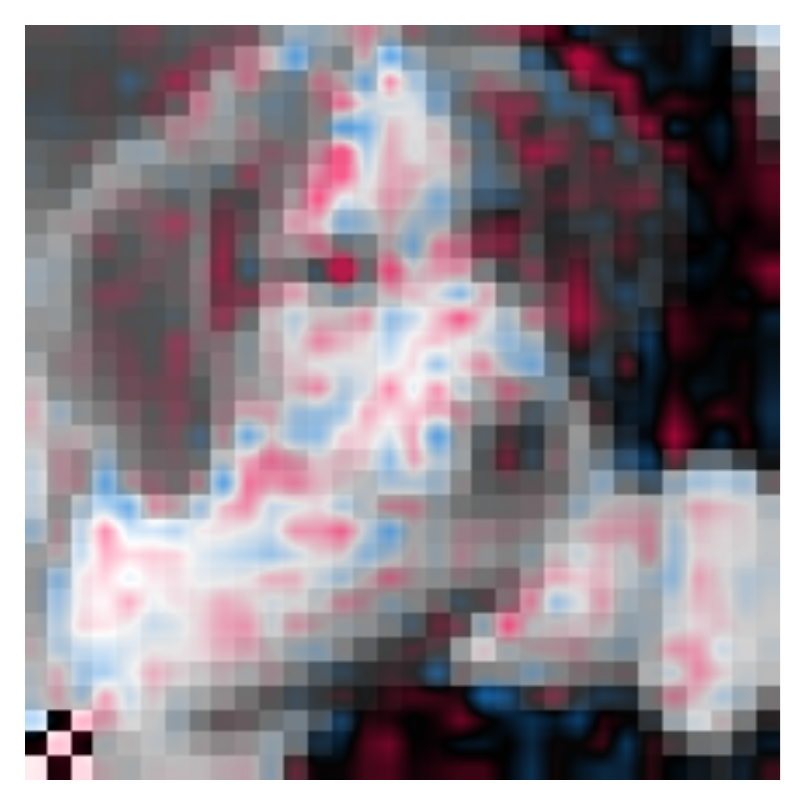} & \shapimg{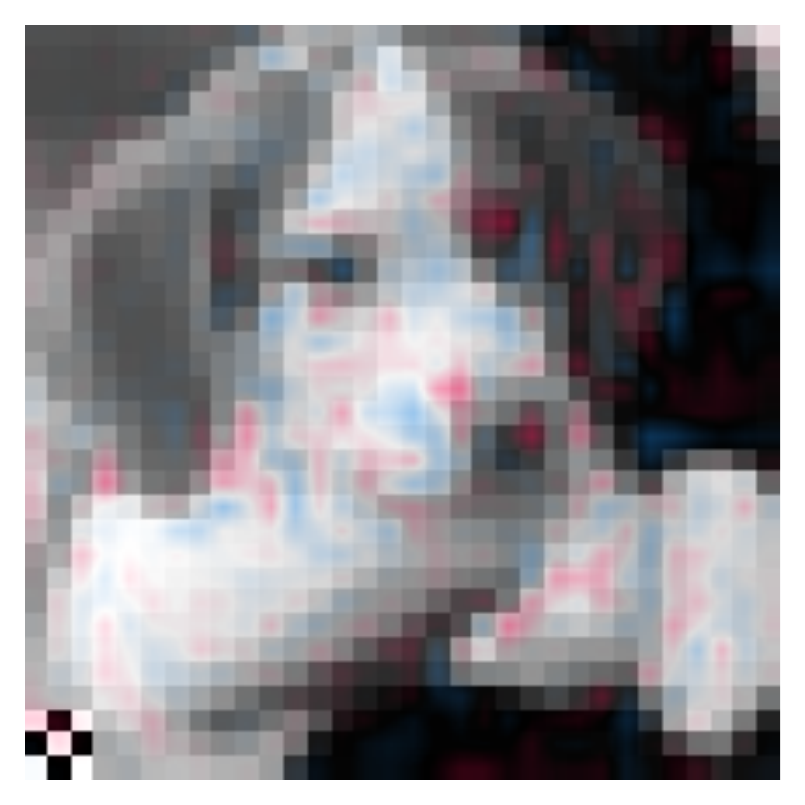} & \shapimg{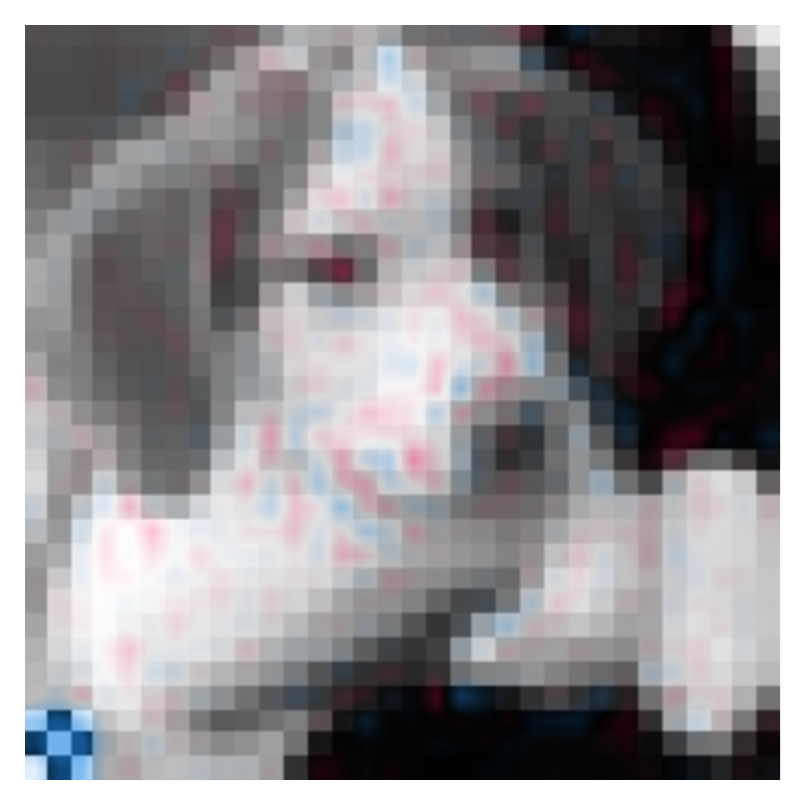} & \shapimg{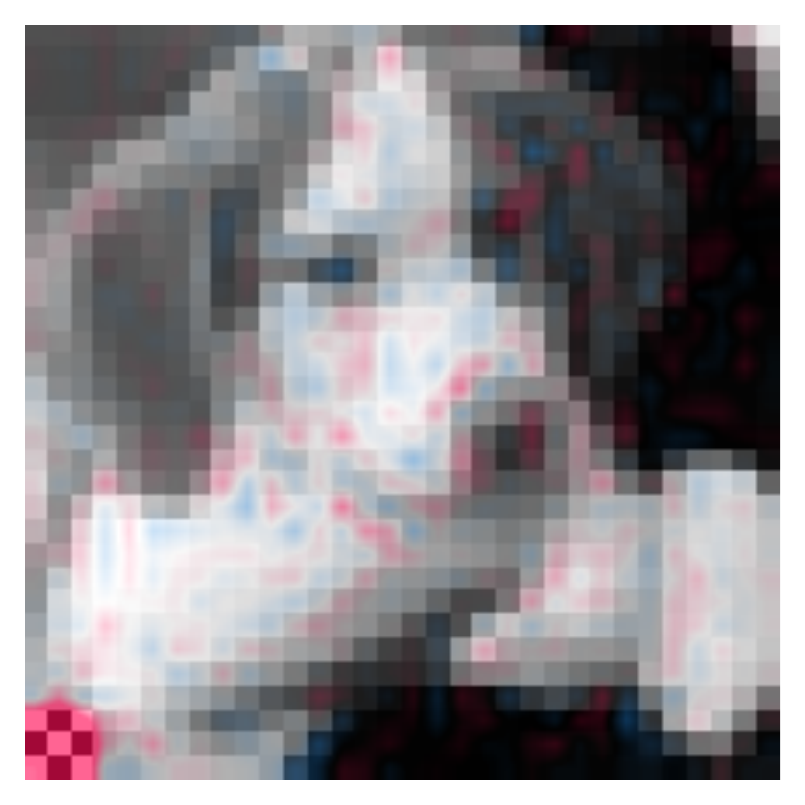} & \shapimg{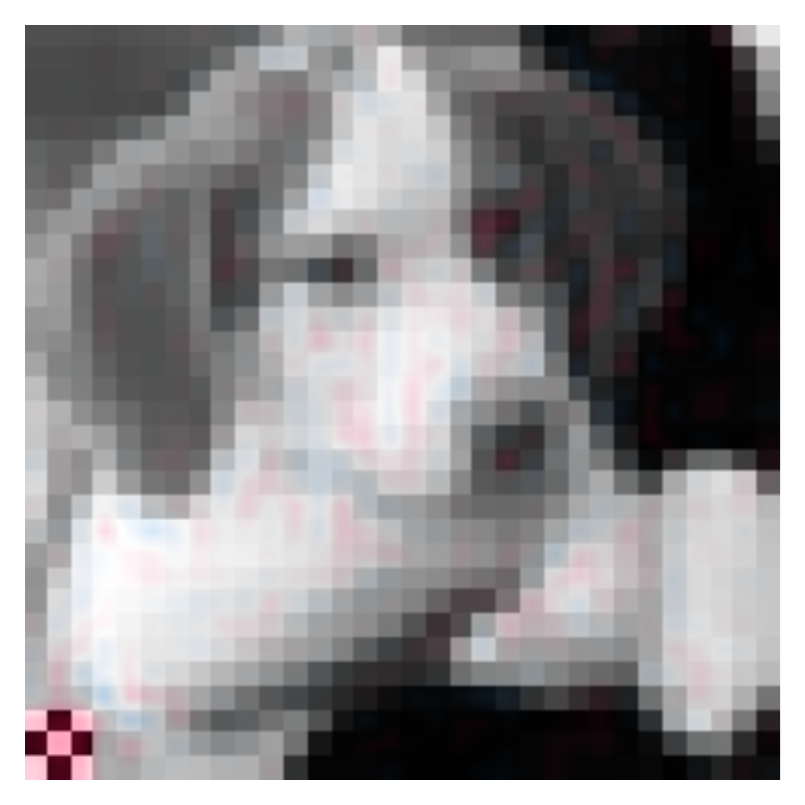} & \shapimg{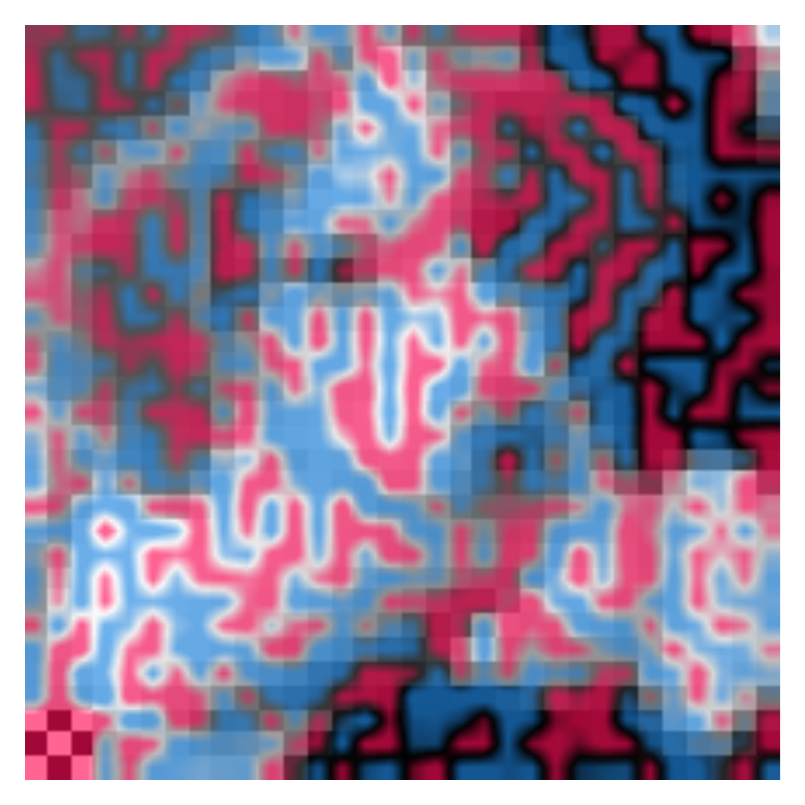} & \shapimg{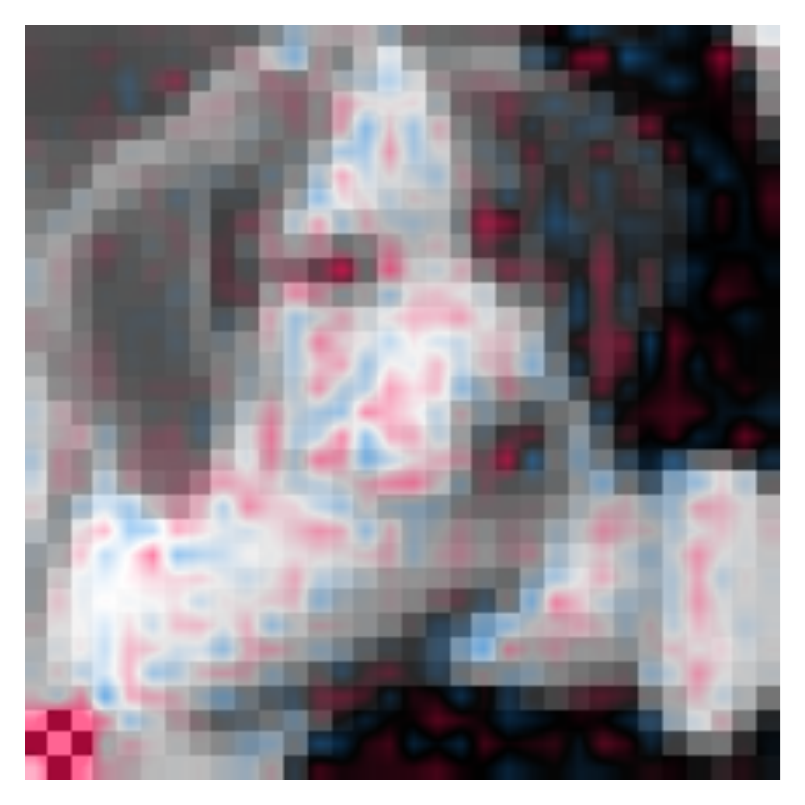} & \shapimg{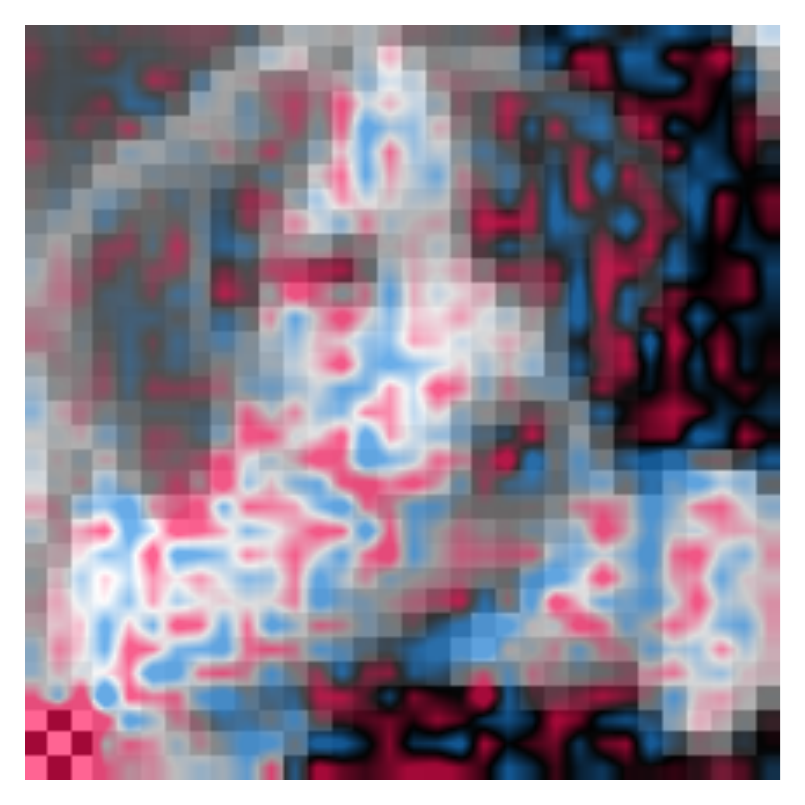} \\\bottomrule
    \end{tabular}
    \caption{SHAP values for the three attacks (and a control), for both the correct Dog class and the backdoored Frog class, with and without the trigger (bottom left). Pixels in \colorbox{shapred}{red\hspace{1pt}} denote a \textbf{positive} contribution to that class, and pixels in \colorbox{shapblue}{blue\hspace{1pt}} denote a \textbf{negative} contribution.}
    \label{fig:shap}
\end{figure}

\section{Datasets}

We use four datasets in our evaluation. The CIFAR-10 dataset \citep{cifar10} contains 50,000 32x32 color training images and 10,000 testing images from 10 common classes; we use this standard dataset unchanged.

For our experiments in Setting 2, we construct a baseline transfer learning setup using the German Traffic Sign Recognition Benchmark (GTSB) \citep{gtsb} as an initial dataset. Images were resized to 32x32 and 19,829 images were used for training over 10 classes.

\begin{figure}[h]
    \centering
    \begin{subfigure}{\linewidth}
        \includegraphics[width=\textwidth]{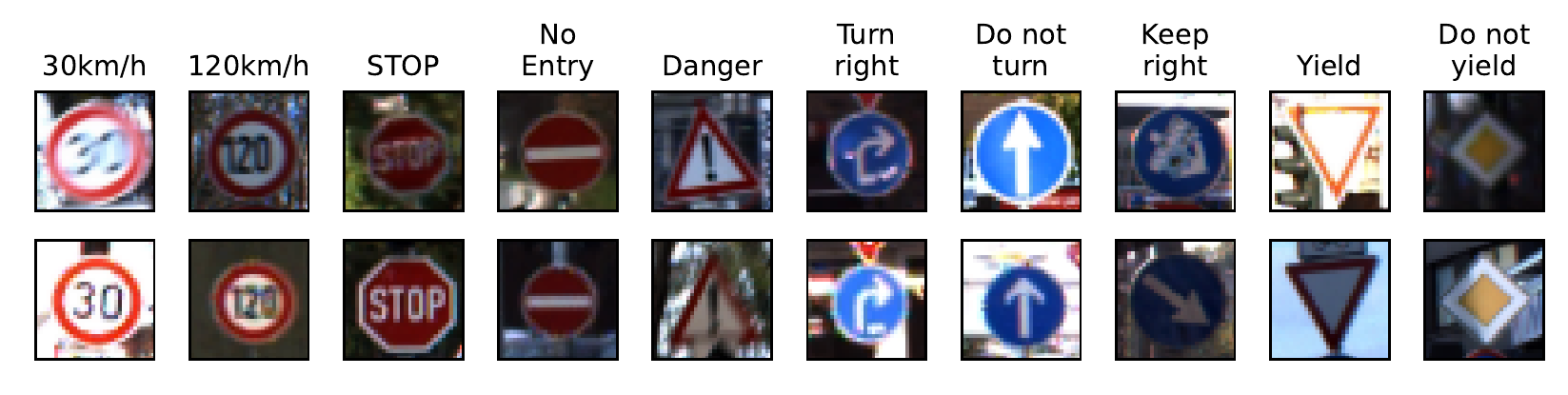}
        \vspace{-18pt}
        \caption{The GTSB dataset}
    \end{subfigure}
    \begin{subfigure}{\linewidth}
        \includegraphics[width=\textwidth]{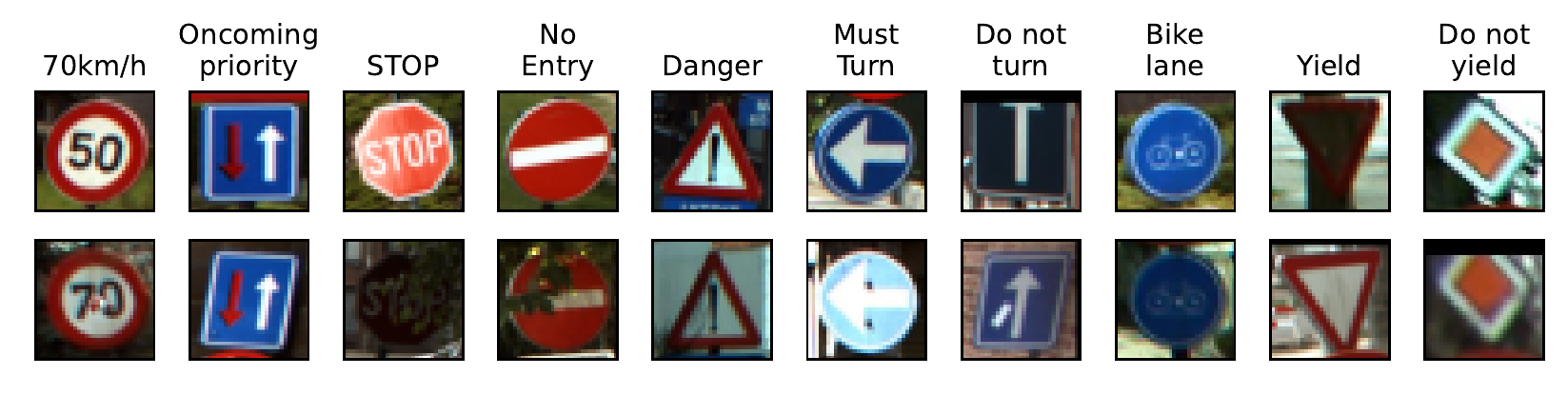}
        \vspace{-18pt}
        \caption{The BTSC dataset}
    \end{subfigure}
    \caption{The correspondence between classes in our fine-tuning datasets, which allows for effective transfer learning.}
    \label{fig:gtsb-btsc}
\end{figure}

The same preprocessing is applied to the Belgian Traffic Sign Classification dataset (BTSC) \citep{btsc} to provide the target dataset for transfer learning (fine-tuning). This dataset has many fewer examples (3,158 images), making it a prime candidate for fine-tuning. 10 classes were selected from both datasets that (a) have a significant number of training examples in GTSB and (b) align between the two datasets, allowing for better transfer learning. Figure \ref{fig:gtsb-btsc} shows the class alignment. The problem of traffic sign detection was motivated by autonomous driving models, as discussed in \cite{gu2017badnets}.

\begin{figure}[h]
    \centering
        \includegraphics[width=\textwidth]{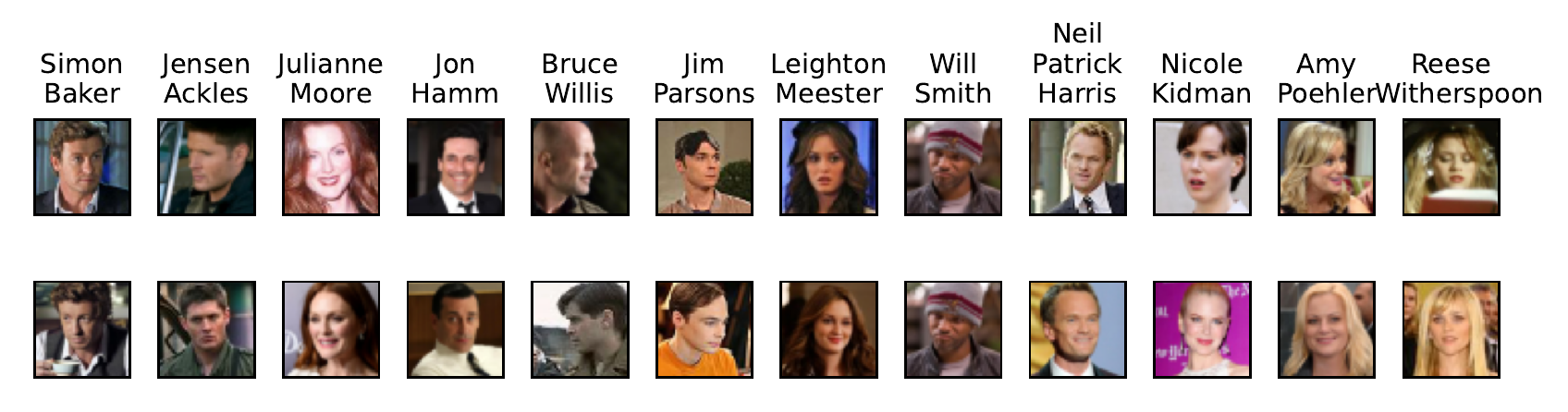}
    \caption{Examples from the IMDB-Wiki face recognition dataset.}
    \label{fig:imdb}
\end{figure}

In Setting 3, we use CIFAR-10 and GTSB in addition to a face classification dataset; motivated by safety-critical applications that an attacker might want to target such as CCTV face detection. The dataset used is IMDB-WIKI \citep{imdbwiki}, where faces are cropped using the provided bounding boxes and images are resized to 48x48. Due to the huge number of classes and large class imbalance, 12 of the most common celebrities were selected as our classes, seen in Figure \ref{fig:imdb}. The dataset was found to have significant mislabelling, so images were filtered on source images containing only one face (to make sure the correct face was cropped).

\section{Licensing}
\label{appendix:license}
The vast majority of the work was implemented ourselves and will be released under the permissive \textbf{MIT license}, which allows future researchers to build on the work unconstrained (only requiring preservation of the license file). All dependencies of our library are similarly released under OSI\footnote{https://opensource.org/licenses}-approved licenses, allowing them all to be easily compiled and installed.

\section{Computational resources}
All experiments complete in $<7$ GPU-days on a single NVIDIA 1080Ti system with a Ryzen Threadripper 2970WX.